\documentclass[twoside,11pt]{article}

\usepackage{jmlr2e,times}

 \usepackage[colorlinks,
            linkcolor=blue,
            citecolor=blue,
            urlcolor=magenta,
            linktocpage,
            plainpages=false]{hyperref}

\usepackage{amssymb,amsmath,color}
\usepackage{url,color}

     \usepackage[mathcal]{eucal}

\title{On the Equivalence between  Kernel Quadrature Rules\\ and Random Feature Expansions}

\author{\name Francis Bach  \email francis.bach@ens.fr \\
\addr  INRIA \\
D\'epartement d'Informatique de l'Ecole Normale Sup\'erieure \\
Paris, France \\
 }

  \usepackage{natbib,color}

 \newcommand{\BEAS}{\begin{eqnarray*}}
\newcommand{\EEAS}{\end{eqnarray*}}
\newcommand{\BEA}{\begin{eqnarray}}
\newcommand{\EEA}{\end{eqnarray}}
\newcommand{\BEQ}{\begin{equation}}
\newcommand{\EEQ}{\end{equation}}
\newcommand{\BIT}{\begin{itemize}}
\newcommand{\EIT}{\end{itemize}}
\newcommand{\BNUM}{\begin{enumerate}}
\newcommand{\ENUM}{\end{enumerate}}
\newcommand{\BA}{\begin{array}}
\newcommand{\EA}{\end{array}}

\newcommand{\tr}{\mathop{ \rm tr}}

\newcommand{\idm}{I}
\newcommand{\rb}{\mathbb{R}}

\newcommand{\mysec}[1]{Section~\ref{sec:#1}}
\newcommand{\eq}[1]{Eq.~(\ref{eq:#1})}
\newcommand{\myfig}[1]{Figure~\ref{fig:#1}}

\def \E{{\mathbb E}}
\def \P{{\mathbb P}}

\def \E{{\mathbb E}}
\def \P{{\mathbb P}}
\def \F{{\mathcal F}}

\def \H{{\mathcal H}}
\def \X{{\mathcal X}}
\def \V{{\mathcal V}}

 \editor{}

\begin{document}

 \maketitle

\begin{abstract}

We show that  kernel-based quadrature rules for computing integrals can be seen as a special case of random feature expansions for positive definite kernels, for a particular decomposition that always exists for such kernels.
 We provide  a theoretical analysis of the number of required samples for a given approximation error, leading to both upper and lower bounds that are based solely on the eigenvalues of the associated integral operator and match up to logarithmic terms.
In particular, we show that   the upper bound may be obtained from independent and identically distributed samples from a specific non-uniform   distribution, while the lower bound if valid for any set of points.
  Applying our results to  kernel-based quadrature, while our results are fairly general, we recover known upper and lower bounds for the special cases of Sobolev spaces. Moreover, our results extend to the more general problem of full function approximations (beyond simply computing an integral), with results in $L_2$- and $L_\infty$-norm that match known results for special cases.
 Applying our results to random features, we show an improvement of the number of random features needed to preserve the generalization guarantees for learning with Lipshitz-continuous losses.

\end{abstract}

\section{Introduction}

The numerical computation of high-dimensional integrals is one of the core computational tasks in many areas of machine learning, signal processing and more generally   applied mathematics, in particular in the context of Bayesian inference~\citep{gelman2004bayesian}, or the study of complex systems~\citep{robert2005monte}. In this paper, we focus on \emph{quadrature rules}, that aim at approximating the  integral of a certain function from only  the (potentially noisy) knowledge of the function values at as few as possible   well-chosen points. Key situations that remain active areas of research are problems where the measurable space where the function is defined on  is either high-dimensional or structured (e.g., presence of discrete structures, or graphs). For these problems, techniques based on \emph{positive definite kernels} have emerged as having the potential to efficiently deal with these situations, and to improve over plain Monte-Carlo integration~\citep{o1991bayes,rasmussen2003bayesian,huszaroptimally,oates2015variance}. In particular, the quadrature problem may be cast as the one of approximating a fixed  element, the mean element~\citep{smola2007hilbert}, of a Hilbert space as a linear combination of well chosen elements, the goal being to minimize the number of these factors as it corresponds to the required number of function evaluations.

A seemingly unrelated problem on positive definite kernels have recently emerged, namely the representation of the corresponding infinite-dimensional  feature space from \emph{random sets of features}. If a certain positive definite kernel between two points may be represented as the expectation of the product of two random one-dimensional (typically non-linear) features computed on these two points, the full kernel (and hence its feature space) may be approximated by sufficiently many random samples, replacing the expectation by a sample average~\citep{neal1995bayesian,rahimi2007random,huang2006extreme}. When using these random features, the complexity of a regular kernel method such as the support vector machine or ridge regression goes from quadratic in the number of observations to linear in the number of observations, with a constant  proportional to the number of random features, which thus drives the running time complexity of these methods.

In this paper, we make the following contributions:

\vspace*{-.25cm}

\begin{list}{\labelitemi}{\leftmargin=1.7em}
   \addtolength{\itemsep}{-.215\baselineskip}

\item[--] After describing the functional analysis framework our analysis is based on and presenting many examples in \mysec{fr}, we show in \mysec{quadd} that these two problems are strongly related; more precisely, optimizing weights in kernel-based quadrature rules can be seen as decomposing a certain function in a special class of random features for a particular decomposition that always exists for all positive definite kernels on a measurable space.

\item[--] We provide in \mysec{bounds} a theoretical analysis of the number of required samples for a given approximation error,  leading to both upper and lower bounds that are based solely on the eigenvalues of the associated integral operator and match up to logarithmic terms.
In particular, we show that  the upper bound may be obtained as independent and identically distributed samples from a specific non-uniform   distribution, while the lower bound if valid for any set of points.

\item[--] Applying our results to  kernel quadrature, while our results are fairly general, we recover known upper and lower bounds for the special cases of Sobolev spaces (\mysec{quad}). Moreover, our results extend to the more general problem of full function approximations (beyond simply computing an integral), with results in $L_2$- and $L_\infty$-norm that match known results for special cases (\mysec{extensions}).

\item[--] Applying our results to random feature expansions, we show in \mysec{consequences} an improvement of the number of random features needed for preserving the generalization guarantees for learning with Lipshitz-continuous losses.

\end{list}

\section{Random Feature Expansions of Positive Definite Kernels}
\label{sec:fr}

Throughout this paper, we consider a  topological space {$\X$} equipped with a  {Borel probability measure~$d \rho$}, which we assume to have full support. This naturally defines the space of square-integrable functions\footnote{For simplicity and following most of the literature on kernel methods,  we identify functions and their equivalence classes for the equivalence relationship of being equal except for a zero-measure (for $d\rho$)  subset of $\X$.}.

\subsection{Reproducing kernel Hilbert spaces and integral operators}
\label{sec:rkhs}

We consider a continuous positive definite kernel $k: \X \times \X \to \rb$, that is a symmetric function such that for all finite families of points in $\X$, the matrix of pairwise kernel evaluations is positive semi-definite.  This thus defines a reproducing kernel Hilbert space (RKHS) $\F$ of functions from $\X$ to $\rb$, which we also assume separable. This RKHS has two important characteristic properties~\citep[see, e.g.,][]{berlinet2004reproducing}:
\BIT
\item[(a)] \emph{Membership of kernel evaluations}: for any $x \in \X$, the function $k(\cdot,x) : y \mapsto k(y,x)$ is an element of~$\F$.
\item[(b)] \emph{Reproducing property}: for all $f \in \F$ and $x \in \X$, $f(x) = \langle f, k(\cdot,x) \rangle_\F$. In other words, function evaluations are equal to dot-products with a specific element of $\F$.
\EIT
Moreover, throughout the paper, we assume that the function $x \mapsto k(x,x)$ is integrable with respect to $d\rho$ (which is weaker than $\sup_{x \in \mathcal{X}} k(x,x) < \infty$). This implies that $\F$ is a subset of $L_2(d\rho)$;  that is, functions in the RKHS $\F$ are all square-integrable for $d\rho$. In general, $\F$ is strictly included in $L_2(d\rho)$, but, in this paper, we will always assume that it is \emph{dense} in $L_2(d\rho)$, that is, any function in $L_2(d\rho)$ may be approximated arbitrarily closely by a function in $\F$. Finally, for simplicity of our notation (to make sure that the sequence of eigenvalues of integral operators is infinite) we will always assume that $L_2(d\rho)$ is infinite-dimensional, which excludes finite sets for $\X$. Note that the last two assumptions (denseness and infinite dimensionality) can easily be relaxed.

\paragraph{Integral operator.} 
Reproducing kernel Hilbert spaces are often studied through a specific integral operator which leads to an isometry with $L_2(d\rho)$~\citep{Sma2001mathematical}. Let $\Sigma:L_2(d\rho) \to L_2(d\rho)$ be    defined as 
$$\displaystyle (\Sigma f) (x) = \int_\X f(y) k(x,y) d\rho(y).$$
Since $\int_\X k(x,x)d\rho(x) $ is finite,  $\Sigma$ is self-adjoint,  positive semi-definite and trace-class~\citep{simon1979trace}. Given that $\Sigma f$ is a linear combination of kernel functions $k(\cdot,y)$, it belongs to $\F$. More precisely,  since we have assumed that $\F$ is dense in $L_2(d\rho)$, $\Sigma^{1/2}$, which is the unique positive self-adjoint square root of $\Sigma$, is a bijection from  $L_2(d\rho)$ to our RKHS $\F$; that is, for any $f \in \F$, there exists a unique $g \in L_2(d\rho)$ such that $f  = \Sigma^{1/2} g$ and $\| f\|_\H =   \|g\|_{L_2(d\rho)}$~\citep{Sma2001mathematical}. This justifies the notation $\Sigma^{-1/2} f$ for $ f \in \F$ and means that $\Sigma^{1/2}$ is an isometry from $ L_2(d\rho)$ to $\F$; in other words, for any functions $f$ and $g$ in $\F$, we have:
$$
\langle f,g \rangle_\F = \langle
\Sigma^{-1/2} f, \Sigma^{-1/2} g
\rangle_{L_2(d\rho)}.
$$
This justifies the view of $\F$ as the subspace of functions $f \in L_2(d\rho)$ such that $\|\Sigma^{-1/2} f\|_{L_2(d\rho)}^2$.
This relationship is even more transparent when considering a spectral decomposition of $\Sigma$.

\paragraph{Mercer decomposition.}  From extensions of Mercer's theorem~\citep{mercer}, there exists an orthonormal \emph{basis} $(e_m)_{m \geqslant 1}$ of $L_2(d\rho)$ and a summable non-increasing sequence of strictly positive eigenvalues $(\mu_m)_{m \geqslant 1}$ such that $\Sigma e_m = \mu_m e_m$.  Note that since we have assumed that $\F$ is dense in $L_2(d\rho)$,  there are no zero eigenvalues. 

Since $\Sigma^{1/2}$ is an isometry from $L_2(d\rho)$ to  $\F$, $(\mu_m^{1/2} e_m)_{m \geqslant 1 }$ is an orthonormal basis of $\F$. Moreover, we can use the eigendecomposition to characterize elements of $\F$ as the functions in $L_2(d\rho)$ such that 
$$
\|\Sigma^{-1/2} f\|_{L_2(d\rho)}^2 =  \sum_{m \geqslant 1} \mu_m^{-1} \langle f, e_m \rangle^2_{L_2(d\rho)}$$
is finite. In other words, once projected in the orthonormal basis $(e_m)_{m \geqslant 1}$, elements $f$ of $\F$ correspond to a certain decay of its decomposition coefficients $( \langle f, e_m \rangle_{L_2(d\rho)}) _{m \geqslant 1}$.

Finally, by decomposing the function $k(\cdot,y): x\mapsto k(x,y)$, we obtain the Mercer decomposition:
$$k(x,y) = \sum_{m \geqslant 1} \mu_m e_m(x) e_m(y).$$

\paragraph{Properties of the spectrum.} The sequence of eigenvalues $(\mu_m)_{m \geqslant 1}$ is an important quantity that appears in the analysis of kernel methods~\citep{hastie_GAM,Cap_DeV:2007,Har_Bac_Mou:2008,bac2012sharp,alaoui2014fast}.
It depends both on the kernel $k$ and the chosen distribution $d \rho$. 

Some modifications of the kernel $k$ or the distribution $d \rho$ lead to simple behaviors for the spectrum.
For example, if we have a second distribution so that $\frac{d\rho'}{d\rho}$ is upper-bounded by a constant $c$, then, as a consequence of the Courant-Fischer minimax theorem~\citep{horn2012matrix}, the eigenvalues for $d \rho'$ are less than than $c$ times that the ones for $d \rho$. Similarly, if the kernel $k'$ is such that $c k - k'$ is a positive definite kernel, then we have a similar bound between eigenvalues. 

In this paper, for any strictly positive $\lambda$, we will also consider the quantity $m^\ast(\lambda)$ equal to the number of eigenvalues $\mu_m$ that are greater than or equal to $\lambda$.  Since we have assume that the sequence $m$ is non-increasing, we have $m^\ast(\lambda) = \max \{ m \geqslant 1, \ \mu_m \geqslant \lambda \}$.
This is a left-continuous non-increasing function, that tends to $+\infty$ when $\lambda$ tends to zero (since we have assumed that there are infinitely many strictly positive eigenvalues), and characterizes the sequence $(\mu_m)_{m \geqslant 1}$, as we can recover $\mu_m$ as $\mu_m = \sup \{ \lambda \geqslant 0, \ m^\ast(\lambda) \geqslant m \}$.

\paragraph{Potential confusion with covariance operator.}  
Note that the operator $\Sigma$ is a  self-adjoint operator on $L_2(d\rho)$. It should not be confused with the (non-centered) covariance operator $C$~\citep{baker1973joint}, which is a self-adjoint operator on a different space, namely the RKHS $\F$, defined by
$  \langle g,C f \rangle_\F= \int_{\X} f(x) g(x) d\rho(x)
 $.
Given that $\Sigma^{1/2}$ is an isometry from $L_2(d\rho)$ to $\F$, the operator $C$ may also be used to define  an operator on $L_2(d\rho)$, which happens to be exactly $\Sigma$. Thus, the two operators have the same eigenvalues. Moreover, we have, for any $y \in \X$:
$$
(Cf)(y) = \langle k(\cdot,y), Cf \rangle_\F = \int_\X k(x,y) f(x) d \rho(x) = (\Sigma f)(y),
$$
that is, $C$ is equal to the restriction of $\Sigma$ on $\F$.

\subsection{Kernels as expectations}
\label{sec:kexp}

On top of the generic assumptions made above, we   assume that there is another
measurable set~$\V$ equipped with a   probability measure $d\tau$. We consider  a function $\varphi: \V \times \X \to \rb$ which is square-integrable (for the measure $d\tau \otimes d\rho$), and  assume that the kernel~$k$ may be written as, for all $x,y \in \X$:
\BEQ
\label{eq:randomfeat}
k(x,y) = \int_\V   \varphi(v,x)  \varphi(v,y)  d\tau(v)
= \langle
\varphi(\cdot,x), \varphi(\cdot,y)
\rangle_{L_2(d\tau)}
.
\EEQ
In other words, the kernel between $x$ and $y$ is simply the expectation of  $\varphi(v,x)  \varphi(v,y)$ for $v$ following the probability distribution $d\tau$.  In this paper, we see $x \mapsto \varphi(v,x) \in \rb$ as a one-dimensional random feature and $\varphi(v,x)  \varphi(v,y)$ is the dot-product associated with this random feature. We could consider extensions where $\varphi(v,x)$ has values in a Hilbert space (and not simply $\rb$), but this is outside the scope of this paper.

\paragraph{Square-root of integral operator.}
Such additional structure allows to give an explicit characterization of the RKHS $\F$ in terms of the features $\varphi$. In terms of operators, the function $\varphi$ leads to a specific square-root of the integral operator $\Sigma$ defined in \mysec{rkhs} (which is not the positive self-adjoint square-root $\Sigma^{1/2}$).

We consider the bounded linear operator $T: L_2(d\tau) \to L_2(d\rho)$ defined as
\BEQ
\label{eq:T}
(Tg)(x) = \int_{\V} g(v) \varphi(v,x) d\tau(v) = \langle g, \varphi(\cdot,x) \rangle_{L_2(d\tau)}.
\EEQ

Given $T: L_2(d\tau) \to L_2(d\rho)$, the adjoint operator
 $T^\ast: L_2(d \rho) \to L_2(d\tau)$ is the unique operator such that
 $ \langle g, T^\ast f \rangle_{L_2(d\tau)} = \langle T g,   f \rangle_{L_2(d\rho)}$ for all $f,g$. Given the definition of $T$ in \eq{T},
 we simply inverse the role of $\V$ and~$\X$ and have:
 $$
 (T^\ast f)(v) = \int_\X f(x) \varphi(v,x) d \rho(x).
 $$
 This implies by Fubini's theorem that 
 \BEAS
 (TT^\ast f) (y) & = &  \int_\V \bigg(\int_\X f(x)  \varphi(v,y) d \rho(x) \bigg) \varphi(v,x) d\tau(v)
\\
& = &   \int_\X f(x)  \bigg( \int_\V  \varphi(v,y)\varphi(v,x) d\tau(v)\bigg)   d \rho(x) 
=   \int_\X f(x)  k(x,y)   d \rho(x)  = (\Sigma f)(y),
 \EEAS
 that is we have an expression of the integral operator $\Sigma$ as $ \Sigma = TT^\ast$. Thus, the decomposition of the kernel $k$ as an expectation corresponds to a particular \emph{square root} $T$ of the integral operator---there are many possible choices for such square roots, and thus many possible expansions like \eq{randomfeat}. It turns out that the positive self-adjoint square root $\Sigma^{1/2}$ will correspond to the equivalence with quadrature rules (see \mysec{reformulation}).

\paragraph{Decomposition of functions in $\F$.} Since $\Sigma = TT^\ast$ and $\Sigma^{1/2}$ is an isometry between $L_2(d\rho)$ and~$\F$, we can naturally expressed any elements of $\F$ through the operator $T$ and thus the features~$\varphi$.

As a linear operator, $T$ defines a bijection from the orthogonal of its null space $ ({\rm Ker}\  T)^\perp \subset L_2(d\tau)$ to its image ${\rm Im}(T) \subset L_2(d\rho)$, and this allows to define uniquely $T^{-1} f \in ({\rm Ker}\  T)^\perp $ for any $f \in {\rm Im}(T)$, and a dot-product on $  {\rm Im}(T)$ as
$$\langle f,h \rangle_{{\rm Im}(T)} = \langle T^{-1} f, T^{-1} g \rangle_{L_2(d\tau)}.$$
As shown by~\citet[App.~A]{relu}, ${\rm Im}(T)$ turns out to be equal to our RKHS\footnote{The proof goes as follows: (a) for any $y \in \X$, $k(\cdot,y)$ can be expressed as $\int_{\V} \varphi(v,y) \varphi(v,\cdot) d\tau(v) = T \varphi(\cdot,y)$ and thus belongs to ${\rm Im}(T)$; (b) for any $f \in {\rm Im}(T)$, and $y \in \X$, we have
$
\langle f, k(\cdot,y) \rangle_{{\rm Im}(T)} = \langle T^{-1} f , \varphi(\cdot,y) \rangle_{L_2(d\tau)} = (TT^{-1} f)(y) = f(y)
$, that is, the reproducing property is satisfied. These two properties are characteristic of $\F$.}. Thus, the norm $\| f \|_\F^2$ for $f \in \F$ is equal to the squared $L_2$-norm of $T^{-1} f \in ({\rm Ker}\  T)^\perp$, which is itself equal to the minimum of $\|g \|^2_{L_2(d\tau)}$ over all $g$ such that $Tg = f$. The resulting $g$ may also be defined through pseudo-inverses.

In other words, a function $f \in L_2(d\rho)$ is in $\F$ if and only if it may be written as
$$ \displaystyle
\forall x \in \X, \ f(x) = \int_\V   g(v)   \varphi(v,x)  d\tau(v) = \langle g, \varphi(\cdot,x) \rangle_{L_2(d\tau)},$$
for a certain function $g: \V \to \rb$
such that $\| g\|^2_{L_2(d\tau)}$ is finite, with a norm $\| f\|_\F^2$ equal to the minimum (which is always attained) of $  \| g\|^2_{L_2(d\tau)},
$
over all possible decompositions of $f$.

 \paragraph{Singular value decomposition.}
 The operator $T$ is an Hilbert-Schmidt operator, to which the singular value decopomposition can be applied~\citep{kato2012perturbation}. That is, there exists an orthonormal basis $(f_m)_{m \geqslant 1}$ of  $ ({\rm Ker}\  T)^\perp \subset L_2(d\tau)$, together with the orthonormal basis $(e_m)_{m \geqslant 1}$ of $L_2(d\rho)$ which we have from the eigenvalue decomposition of $\Sigma = TT^\ast$,  such that $T f_m = \mu_m^{1/2} e_m$. Moreover, we have:
 \BEQ
\label{eq:phisvd} \varphi(v,x) = \sum_{m \geqslant 1} \mu_m^{1/2} e_m(x) f_m(v),
 \EEQ
 with a convergence in $L_2( d\tau \otimes d\rho)$. This extends the Mercer decomposition of the kernel $k(x,y)$.
    
 \paragraph{Integral operator as an expectation.} Given the expansion of the kernel $k$ in \eq{randomfeat}, we may express the integral operator $\Sigma$ as follows, explicitly as an expectation:
 \BEA
\nonumber \Sigma f & = & \int_\X f(y) k(\cdot,y) d\rho(y)
 = \int_\X  \int_\V f(y) \varphi(v,\cdot) \varphi(v,y) d \rho(y) d\tau(v) \\
\label{eq:sigmaphi} & = & \int_\V \varphi(v,\cdot) \langle \varphi(v,\cdot), f \rangle_{L_2(d \rho)} d\tau(v)= 
 \bigg(
  \int_\V \varphi(v,\cdot) \otimes_{L_2(d \rho)}  \varphi(v,\cdot) d\tau(v)
  \bigg) f,
 \EEA
 where $a \otimes_{L_2(d \rho)} b$ is the  operator $L_2(d \rho) \to L_2(d \rho)$ so that 
 $(a \otimes_{L_2(d \rho)} b) f = \langle b,f \rangle_{ L_2(d \rho)} a$.
 This will be useful to define empirical versions, where the integral over $d\tau$ will be replaced by a finite average.

\subsection{Examples}
\label{sec:examples}

In this section, we provide examples of kernels and usual decompositions. We first start by decompositions that always exist, then focus on specific kernels based on Fourier components.

\paragraph{Mercer decompositions.}  The Mercer decomposition provides an expansion for all kernels, as follows:
$$ \displaystyle
k(x,y) =  \sum_{m \geqslant 1 } \frac{\mu_m}{\tr \Sigma} 
\Big[\big(\tr \Sigma)^{1/2}  e_m(x) \Big] \! \cdot \! \Big[\big(\tr \Sigma)^{1/2}  e_m(x) \Big],
$$
which can be transformed in to an expectation with $\V = \mathbb{N}^\ast$. In \mysec{reformulation}, we provide another generic decomposition with~$\V = \X$.   Note that this decomposition is typically impossible to compute (except for special cases below, i.e., special pairs of kernels $k$ and distributions $d\rho$).

\paragraph{Periodic kernels on $[0,1]$.}  We consider $\X = [0,1]$ and translation-invariant kernels $k(x,y)$ of the form $k(x,y) = t(x-y)$, where $t$ is a square-integrable 1-periodic function. These kernels are positive definite if and only if the Fourier series of $t$ is non-negative~\citep{wah1990splines}. An orthonormal basis of $L_2([0,1])$ is composed of the constant function $c_0: x \mapsto 1$ and the functions $c_m: x \mapsto \sqrt{2} \cos 2 \pi m x$ and $s_m: x \mapsto \sqrt{2} \sin 2 \pi m x$. A kernel may thus be expressed as
$$
k(x,y) =\nu_0 c_0(x) +  \sum_{m > 0 } \nu_m \big[ c_m(x) c_m(y) + s_m(x) s_m(y) \big]
= \nu_0 + 2 \sum_{m > 0 } \nu_m \cos 2 \pi m (x-y).
$$
 This can be put trivially as an expectation with $\V= \mathbb{Z}$ and leads to the usual Fourier features \citep{rahimi2007random}.
This is also exactly a Mercer decomposition for $k$ and the uniform distribution on $[0,1]$,  with eigenvalues $\nu_0$ and $\nu_m$, $m >0$ (each of these with multiplicity 2). The associated RKHS norm for a function $f$ is then equal to
$$
\| f \|_\F^2 =  \nu_0^{-1} \Big(
\int_0^1 f(x) dx
\Big)^2 + 2 \sum_{m>0} 
 \nu_m^{-1} \bigg[ \Big(
\int_0^1 f(x) \cos 2 \pi m xdx
\Big)^2 + \Big(
\int_0^1 f(x) \sin 2 \pi m xdx
\Big)^2 \bigg].
$$
A particularly interesting example is obtained through derivatives of $f$. If $f$ is differentiable and has a derivative $f' \in L_2([0,1])$, then, on the Fourier series coefficients of $f$, taking the derivative corresponds to multiplying the two $m$-th coefficients by $2 \pi m$ and swapping them. Sobolev spaces for periodic functions on $[0,1]$ (i.e., such that $f(0)\!=\!f(1)$) are defined through integrability of derivatives~\citep{adams2003sobolev}. In the Hilbert space set-up, a function $f$ belongs to the Sobolev space of order $s$ if one can define  a $s$-th order square-integrable derivative in $L_2$ (for the Lebesgue measure, which happens to be equal to $d \rho$), that is, $f^{(s)} \in L_2([0,1])$. The Sobolev squared norm is then defined as any positive linear combination of the quadratic forms $\int_0^1 f^{(t)}(x)^2 dx$, $t \in \{0,\dots,s\}$, with   non-zero coefficients for $t=0$ and $t=s$ (all of these norms are then equivalent). If only using $t=0$ and $t=s$ with non-zero coefficients, we need $\nu_0^{-1} = 1$ and
$  \nu_m^{-1} = 1 + m^{2s}$. An equivalent   (i.e, with upper and lower bounded ratios) sequence is obtained by replacing $\nu_m
= ( 1 + m^{2s} )^{-1}$ by $\nu_m =   m^{-2s}$, leading to a closed-form formula:
$$k(x,y) = 1+\frac{(-1)^{s-1} (2 \pi)^{2s}}{  (2s)!}B_{2s}(\{x-y\}),$$ where $\{x-y\}$ denotes the fractional part of $x-y$, and $B_{2s}$ is the $2s$-th Bernoulli polynomial~\citep{wah1990splines}. The RKHS $\F$ is then the Sobolev space of order $s$ on $[0,1]$, with a norm equivalent to any of the family of Sobolev norms; it will be used as a running example throughout this paper.

\paragraph{Extensions to $[0,1]^d$.}

	In order to extend to $d>1$, we may consider several extensions as described by~\citet{oates2015variance}, and compute the resulting eigenvalues of the integral operators.
		For simplicity, we consider the Sobolev space on $[0,1]$, with $\nu_0 = 1$ and $\nu_m^{-1} =    m^{2s}$ for $m>0$. The first possibility to extend to $[0,1]^d$ is to take a  kernel which is simply the pointwise product of individual kernels on $[0,1]$. That is, if $k(x,y)$ is the kernel on $[0,1]$, define $K(X,Y) = \prod_{j=1}^d k(x_j,y_j)$ between $X$ and $Y$ in $[0,1]^d$. As shown in Appendix~\ref{app:product}, this leads to eigenvalue decays  bounded by $( \log m)^{2s(d-1)} m^{-2s}$, and thus up to logarithmic terms at the same speed $m^{-2s}$ as $d=1$. While this sounds attractive in terms of generalization performance, it corresponds to a space a function which is not a Sobolev  space in $d$ dimensions. That is the associated squared norm on $f$ would be equivalent  to a linear combination of squared $L_2$-norm of partial derivatives
		$$ \int_{[0,1]^d} \Big(
		\frac{\partial^{t_1+\cdots+t_d} f}{\partial x_1^{t_1} \cdots \partial x_d^{t_d} }
		\Big)^2 dx$$
		for all $t_1,\dots,t_d$ in $\{0,\dots,s\}$. This corresponds to functions which have square-integrable partial derivatives with all \emph{individual} orders less than $s$. All values of $s \geqslant 1$ are allowed and lead to an RKHS.

		This is thus to be contrasted with the usual multi-dimensional Sobolev space which is composed of functions which have square-integrable partial derivatives with all orders $(t_1,\dots,t_d)$ with \emph{sum}  $t_1+\cdots + t_d$ less than $s$. Only $s> d/2$ is then allowed to get an RKHS. The Sobolev norm is then of the form		$$
		\sum_{ t_1 + \cdots + t_d  \leqslant s} \int_{[0,1]^d} \Big(
		\frac{\partial^{t_1+\cdots+t_d} f}{\partial x_1^{t_1} \cdots \partial x_d^{t_d} }
		\Big)^2 dx.
		$$
		In the expansion on the $d$-th order tensor product of the Fourier basis, the norm above is equivalent to putting a weight on the element $(m_1,\dots,m_d)$ asymptotically equivalent to $\big( \sum_{j=1}^d m_j \big)^{2s}$, which thus represent the inverse of the eigenvalues of the corresponding kernel for the uniform distribution $d\rho$ (this is simply an explicit Mercer decomposition). Thus, the number of eigenvalues which are greater than $\lambda$ grows as the number of 		$(m_1,\dots,m_d)$  such that their sum is less than $\lambda^{-1/(2s)}$, which itself is less than a constant times 
		$\lambda^{-d/(2s)}$ (see a proof in Appendix~\ref{app:product}). This leads to an eigenvalue decay of $m^{-2s/d}$, which is much worse because of the term in $1/d$ in the exponent.

\paragraph{Translation invariant kernels on $\rb^d$.}  We consider $\X = \rb^d$ and translation-invariant kernels $k(x,y)$ of the form $k(x,y) = t(x-y)$, where $t$ is an integrable function from $\rb^d$ to $\rb$. It is known that these kernels are positive definite if and only if the Fourier transform of $t$ is always a non-negative real number. More precisely,  
if $\hat{t}(\omega) = \int_{\rb^d} t(x) e^{-i \omega^\top  x} dx \in \rb_+$, then 
$$
k(x,y) =  \frac{1}{(2\pi)^d} \int_{\rb^d} \hat{t}(\omega) e^{i \omega^\top (x-y)} d \omega
= \frac{1}{(2\pi)^d}  \int_{\rb^d} \hat{t}(\omega) 
\big[
\cos \omega^\top x 
\cos \omega^\top y + \sin \omega^\top x 
\sin \omega^\top y
\big]
 d \omega.
$$
Following~\citet{rahimi2007random}, by sampling $\omega$ from a density proportional to $\hat{t}(\omega) \in \rb_+$ and $b$ uniformly in $[0,1]$ (and independently of $\omega$), then by defining $\V = \rb^d \times [0,1]$ and $\varphi(\omega,b,x) = \sqrt{2}\cos (\omega^\top x + 2 \pi b)$, we obtain the kernel $k$.

 For these kernels, the decay of eigenvalues has been well-studied by~\citet{Widom}, who relates the decay of eigenvalues to the tails of the distribution $d\rho$ and the decay of the Fourier transform of $t$. For example, for the Gaussian kernel where $k(x,y) = \exp( - \alpha \| x - y \|_2^2)$, on sub-Gaussian distributions, the decay of eigenvalues is geometric, and for kernels leading to Sobolev spaces of order $s$, such as the Matern kernel~\citep{furrer2007framework}, the decay is of the form $m^{-2s/d}$. See also examples by~\citet{birman,Har_Bac_Mou:2008}.

Finally, note that in terms of computation, there are extensions to avoid linear complexity in $d$~\citep{le2013fastfood}.

 \paragraph{Kernels on hyperspheres.}    
If $\X \subset \rb^{d+1}$ is the $d$-dimensional hypersphere $\{ x \in \rb^{d+1}, \| x\|_2^2 = 1\}$, then specific kernels may be used, of the form $k(x,y) = t(x^\top y)$, where $t$ has to have a positive Legendre expansion~\citep{smola2001regularization}. Alternatively, kernels based on neural networks with random weights are directly in the form of random features~\citep{cho2009kernel,relu}: for example,  the kernel $k(x,y) = \E (v^\top x )_+^s   (v^\top x )_+^s $ for $v$ uniformly distributed in  the hypersphere
corresponds to sampling weights in a one-hidden layer neural network with rectified linear units~\citep{cho2009kernel}.
It turns out that these kernels have a known decay for their spectrum. 

As shown by~\citet{smola2001regularization,relu}, the equivalent of Fourier series (which corresponds to $d=1$) is then the basis of spherical harmonics, which is organized by integer frequencies $k \geqslant 1$; instead of having 2 basis vectors (sine and cosine) per frequency, there are $O(k^{d-1})$ of them. As shown by~\citet[page 44]{relu}, we have an explicit expansion of $k(x,y)$ in terms of spherical harmonics, leading to a sequence of eigenvalues equal to $k^{-d-2s-1}$ on the entire subspace associated with frequency~$k$. Thus, by taking multiplicity into account, 
after $\sum_{j=1}^k j^{d-1} \approx k^d$ (up to constants) eigenvalues, we have an eigenvalue of $k^{-d-2s-1}$; this leads to an eigenvalue decay (where all eigenvalues are ordered in decreasing order and we consider the $m$-th one) as   $(m^{1/d})^{ - d - 2s - 1} = m^{-1-1/d - 2s/d}$.

\subsection{Approximation from randomly sampled features}
\label{sec:approxrand}

Given the formulation of $k$ as an expectation in \eq{randomfeat}, it is natural to
consider sampling $n$ elements $v_1,\dots,v_n \in \V$ from the distribution $d\tau$ and  define the kernel approximation 
\BEQ
\label{eq:A}\hat{k}(x,y) = \frac{1}{n} \sum_{i=1}^n 
  \varphi({v_i},x) \varphi({v_i},y) , 
  \EEQ which defines a finite-dimensional RKHS $\hat{\F}$.
  
  From the strong law of large numbers---which can be applied because we have the finite expectation $\E  | \varphi({v},x) \varphi({v},y)|
  \leqslant \big( \E  | \varphi({v},x)|^2  \E  | \varphi({v},y)|^2 \big)^{1/2}$, when $n$ tends to infinity, $\hat{k}(x,y)$ tends to $k(x,y)$ almost surely, and thus we get as tight as desired approximations of the kernel $k$, for a given pair $(x,y) \in \X \times \X$. \citet{rahimi2007random} show that for translation-invariant kernels on a Euclidean space, then the convergence is uniform over  a compact subset of $\X$, with the traditional rate of convergence of $\sqrt{\frac{\log n}{n}}$.    
  
  In this paper, we rather consider \emph{approximations of functions} in $\F$ by functions in $\hat{\F}$, the RKHS associated with $\hat{k}$. A key difficulty is that in general $\hat{\F}$ is not even included in $\mathcal{F}$, and therefore, we cannot use the norm in $\F$ to characterize approximations. In this paper, we choose the $L_2$-norm associated with the probability measure $d\rho$ on $\X$ to characterize the approximation. Given $f \in \F$ with norm $\| f \|_\F$ less than  one, we look for a function $\hat{f} \in \hat{\F}$ of the smallest possible norm and so that $\| f - \hat{f} \|_{L_2(d\rho)}$ is as small as possible.
  
  Note that the measure $d\tau$ is associated to the kernel $k$ and the random features $\varphi$, while the measure~$d\rho$ is associated to the way we want to measure errors (and leads to a specific defintion of the integral operator $\Sigma$).
   
  \paragraph{Computation of error.} 
  Given the definition of the Hilbert space $\F$ in terms of $\varphi$ in \mysec{kexp}, given $g \in L_2(d\tau)$ with   $\| g\|_{L_2(d\tau)} \leqslant 1$
  and $f(x) = \int_\V   g(v)  \varphi(v,x)   d\tau(v)$,
   we aim at finding  an element of $\hat{\F}$ close to $f$. We can also represent $\hat{\F}$ through a similar decomposition, now with a finite number of features, i.e., through 
   $\alpha \in \rb^n$ such that $\hat{f} =  \sum_{i=1}^n   \alpha_i  \varphi(v_i,\cdot)  $ with norm\footnote{Note the factor $n$ because our finite-dimensional kernel in \eq{A} is an average of kernels and not a sum.}  $\| \hat{f} \|^2_{\hat{\F}}  \leqslant  n \| \alpha\|_2^2     $ as small as possible and so that the following approximation error is also small: 
\BEQ
\label{eq:randomerror}\|  \hat{f} - f \|_{L_2(d\rho)} 
= \bigg\| \sum_{i=1}^n      \alpha_i   \varphi(v_i,\cdot)   - \int_\V   g(v)  \varphi(v,\cdot)   d\tau(v)  \bigg\|_{L_2(d\rho)}.
\EEQ 

Note that with $\alpha_i = \frac{1}{n} g(v_i)$ and $v_i$ sampled from $d\tau$ (independently), then, we have
 $\E ( \| \alpha\|_2^2  ) = \sum_{i=1}^n \E \alpha_i^2 = \frac{1}{n} \E g(v)^2 \leqslant \frac{1}{n}$ and an expected error
$
\E  ( \| f - \hat{f} \|_{L_2(d\rho)}^2  )  
= \frac{1}{n} \E \| g(v) \varphi(v,\cdot)\|_{L_2(d\rho)}^2
\leqslant \frac{1}{n} \sup_{v \in \V}  \| \varphi(v,\cdot)\|_{L_2(d\rho)}^2$; our goal is to obtain an error rate with a better scaling in $n$, by (a) choosing a better distribution than $d\tau$ for the points $v_1,\dots,v_n$ and (b) by finding the best possible weights $\alpha \in \rb^n$ (that should of course depend on the function $g$).

\paragraph{Goals.} 
We thus aim at sampling $n$ points $v_1,\dots,v_n \in \V$ from a distribution with density $q$ with respect to $d\tau$. Then the kernel approximation  using \emph{importance weights} is equal to 
$$\hat{k}(x,y) = \frac{1}{n} \sum_{i=1}^n \frac{1}{q(v_i)}
  \varphi({v_i},x) \varphi({v_i},y) $$ (so that the  law of large numbers leads to an approximation converging to $k$), and we thus aim at minimizing
  $
   \Big\| \sum_{i=1}^n      \frac{\beta_i}{q(v_i)^{1/2}}   \varphi(v_i,\cdot)   - \int_\V   g(v)  \varphi(v,\cdot)   d\tau(v)  \Big\|_{L_2(d\rho)}
   $,
   with $n \| \beta\|_2^2$   (which represents the norm of the approximation in $\hat{\F}$ because of our importance weights are taken into account) as small as possible.

\section{Quadrature in RKHSs}
\label{sec:quadd}

\label{sec:quadRK}
Given a square-integrable (with respect to $d \rho$) function $g: \X \to \rb$, the quadrature problem aims at approximating, for all $h \in \F$, integrals 
$$\int_\X h(x) g(x) d\rho(x)$$ by linear combinations 
$$\sum_{i=1}^n \alpha_i h(x_i)$$ of evaluations $h(x_1),\dots,h(x_n)$ of the function $h$ at well-chosen points $x_1,\dots,x_n \in \X$. Of course, coefficients $\alpha \in \rb^n$ are allowed to depend on $g$ (they will in linear fashion in the next section), but not on $h$, as the so-called quadrature rule has to be applied to all functions in $\F$.

\subsection{Approximation of the mean element}

Following~\citet{smola2007hilbert},  the error may be expressed using the reproducing property as:
$$
\sum_{i=1}^n \alpha_i h(x_i) - \int_\X h(x) g(x) d\rho(x)
= \bigg\langle h, 
\sum_{i=1}^n \alpha_i k(\cdot,x_i) - \int_\X k(\cdot,x) g(x) d \rho(x)
\bigg\rangle_{\!\!\F},
$$
 and by Cauchy-Schwarz inequality its supremum over $\| h\|_\F \leqslant 1$ is equal to
 
\BEQ
\label{eq:quadrkhs}
 \bigg\| 
\sum_{i=1}^n \alpha_i k(\cdot,x_i) - \int_\X k(\cdot,x) g(x) d \rho(x)
\bigg\|_\F.
\EEQ
 
The goal of quadrature rules formulated in a RKHS is thus to find points $x_1,\dots,x_n \in \X$ and weights $\alpha \in \rb^n$ so that the quantity in \eq{quadrkhs} is as small as possible~\citep{smola2007hilbert}. 
For $g=1$, the function $\int_\X k(\cdot,x)  d \rho(x)$ is usually referred to as the mean element of the distribution $d\rho$.

The standard Monte-Carlo solution is to consider $x_1,\dots,x_n$ sampled i.i.d.~from $d\rho$ and the weights $\alpha_i = g(x_i) / n$, which leads to a decrease of the error in $1/\sqrt{n}$, with $\E \| \alpha\|_2^2 \leqslant \frac{1}{n}$ and an expected squared error which is equal to $ \frac{1}{n} \E \| g(v) k(:,x) \|_\F^2 \leqslant \frac{1}{n} \| g \|_{L_2(d \rho)}^2
 \sup_{x \in \X} k(x,x)$~\citep{smola2007hilbert}. Note that when $g=1$, \eq{quadrkhs} corresponds to a particular metric between the distribution $d\rho$ and its corresponding empirical distribution~\citep{sriperumbudur2010hilbert}.

In this paper, we explore sampling points $x_i$ from a probability distribution on $\X$ with density $q$ with respect to $d\rho$. Note that when $g$ is a constant function, it is sometimes required that the coefficients~$\alpha$ are non-negative and sum to a fixed constant (so that constant functions are exactly integrated). We will not pursue this here as our theoretical results do not accommodate such constraints~\citep[see, e.g.,][and references therein]{chensuper,bach2012equivalence}.

\paragraph{Tolerance to noisy function values.} 
In practice, independent (but not necessarily identically distributed) noise $\varepsilon_i$ may be present with variance $\sigma^2(x_i)$. Then, the worst (with respect to $\|h\|_\F\leqslant 1$) expected (with respect to the noise) squared error is 
\BEAS
& &  \inf_{\| h \|_\F \leqslant 1} \E \bigg|
 \sum_{i=1}^n \alpha_i ( h(x_i) + \varepsilon_i) - \int_\X h(x) g(x) d\rho(x)
\bigg|^2\\
& = &  \bigg\| 
\sum_{i=1}^n \alpha_i k(\cdot,x_i) - \int_\X k(\cdot,x) g(x) d \rho(x)
\bigg\|_\F^2 + \sum_{i=1}^n \alpha_i^2 \sigma^2(x_i),
\EEAS
and thus in order to be robust to noise, having a small weighted $\ell_2$-norm for the coefficients $\alpha \in \rb^n$ is important.

\subsection{Reformulation as random features}
\label{sec:reformulation}

 For any $x \in \X$, the function $k(\cdot,x)$ is in $\F$, and since we have assumed that $\Sigma^{1/2}$ is an isometry from 
 $  L_2(d\rho)$ to $\F$, there exists  a unique element, which we denote $\psi(\cdot,x)$, of $  L_2(d\rho)$  such that 
$\Sigma^{1/2} \psi( \cdot,x) = k(\cdot,x)$. Given the Mercer decomposition $k(\cdot,x) = \sum_{m \geqslant 1} \mu_m e_m(x) e_m$, we have the expansion $\psi(\cdot,x) = \sum_{ m \geqslant 1} \mu_m^{1/2} e_m(x) e_m$ (with convergence in the $L_2$-norm for the measure $d\rho \otimes d\rho$; note that we do not assume that $\mu_m^{1/2}$ is summable), and thus we may consider $\psi$ as a symmetric function. Note that $\psi$ may not be easy to compute in many practical cases (except for some periodic kernels on $[0,1]$).

We thus have for $(x,y) \in \X \times \X$:
\BEA
\nonumber
k(x,y) & = &  \langle k(\cdot,x), k( \cdot,y) \rangle_\F 
= \langle \Sigma^{1/2} \psi(\cdot,x),  \Sigma^{1/2} \psi(\cdot,y)\rangle_\F  =  \langle \psi(\cdot,x), \psi(\cdot,y) \rangle_{L_2(d\rho)}  \\
\nonumber& & \hspace*{6cm} \mbox{ because of the isometry property of } \Sigma^{1/2}, \\
& = & 
 \int_\X \psi(v,x) \psi(v,y) d \rho(v).
\EEA
 
That is, the kernel $k$ may always be written as an expectation. Moreover, we have the quadrature error in \eq{quadrkhs} equal to (again using the isometry $\Sigma^{1/2}$ from $L_2(d\rho)$ to $\F$):
\BEAS
\bigg\| 
\sum_{i=1}^n \alpha_i k(\cdot,x_i) - \int_\X k(\cdot,x) g(x) d \rho(x)
\bigg\|_\F
& \!\!\!=\!\!\! & \bigg\| 
\sum_{i=1}^n \alpha_i \Sigma^{1/2} \psi(x_i,\cdot) - \int_\X \Sigma^{1/2} \psi(x,\cdot ) g(x) d \rho(x)
\bigg\|_\F
\\[-.1cm]
& \!\!\!=\!\!\! & \bigg\| 
\sum_{i=1}^n \alpha_i  \psi(x_i,\cdot ) - \int_\X   \psi(x,\cdot) g(x) d \rho(x)
\bigg\|_{L_2(d\rho)},
\EEAS

which is exactly an instance of  the approximation result in \eq{randomerror} with $\V = \X$ and $\varphi = \psi$, that is the random feature is indexed by the same set $\X$ as the kernel. Thus, the quadrature problem, that is finding points $x_i$ and weights $(\alpha_i)$ to get the best possible error over all functions of the unit ball of $\F$, is a \emph{subcase} of the random feature problem for a specific expansion.
Note that this random decomposition in terms of $\psi$ is always possible (although not in closed form in general). 

\paragraph{Interpretation through square-roots of intergral operators.} As shown in \mysec{kexp}, random feature expansions correspond to square-roots of the integral operator $\Sigma: L_2(d\rho) \to L_2(d\rho)$ as $\Sigma = TT^\ast$. Among the many possible square roots, the quadrature case corresponds exactly to the positive self-adjoint square root $T=\Sigma^{1/2}$. In this situation, the basis $(f_m)_{m \geqslant 1}$ of the singular value decomposition of $T=\Sigma^{1/2}$ is equal to $(e_m)_{m \geqslant 1}$, recovering the expansion $\psi(x,y) = \sum_{m \geqslant 1} \mu_m^{1/2} e_m(x) e_m(y)$ which we have seen above.

\paragraph{Translation-invariant kernels on $[0,1]^d$ or $\X = \rb^d$.} In this important situation, we have two different expansions: the one based on Fourier features, where the random variable indexing the one-dimensional feature is a \emph{frequency}, while for the one based on the square root $\psi$, the random variable is a \emph{spatial variable} in $\X$. As we show in \mysec{bounds}, our results are independent of the chosen expansions and thus apply to both. However, (a) when the goal is to do quadrature, we need to use~$\psi$, and (b) in general, the decomposition based on Fourier features can be easily computed once samples are obtained, while for most kernels, $\psi(x,y)$ does not have any closed-form simple expression. In \mysec{simu}, we provide a simple example with $\X = [0,1]$ where the two decompositions are considered.

\paragraph{Goals.}  In order to be able to make the parallel with random feature approximations, we consider importance-weighted coefficients $\beta_i = \alpha_i q(x_i)^{1/2}$, and we thus aim at minimizing  the approximation error
$$
\Big\| 
\sum_{i=1}^n \beta_i q(x_i)^{-1/2} k(\cdot,x_i) - \int_\X k(\cdot,x) g(x) d \rho(x)
\Big\|_\F.
$$
We consider potential independent noise with variance $\sigma^2(x_i) \leqslant \tau^2 q(x_i)$ for all $x_i$, so that the tolerance to noise is characterized by the $\ell_2$-norm $\|\beta\|_2$.

\subsection{Relationship with column sampling}
The problem of quadrature is related to the problem of column sampling. Given $n$ observations $x_1,\dots,x_n \in \X$, the goal of column-sampling methods is to approximate the $n \times n$ matrix of pairwise kernel evalulations, the so-called \emph{kernel matrix}, from a subset of its columns.  It has appeared under many names: Nystr\"om method~\citep{williams2001using}, sparse greedy approximations~\citep{smola2000sparse}, incomplete Cholesky decomposition~\citep{fine01efficient}, Gram-Schmidt orthonormalization~\citep{Cristianini2004} or CUR matrix decompositions~\citep{mahoney2009cur}.

While column sampling has typically been analyzed for a fixed kernel matrix, it has a natural extension which is related to  quadrature problems: selecting $n$ points $x_1,\dots,x_n$ from $\X$ such that the projection of any element of the RKHS $\F$ onto the subspace spanned by $k(\cdot,x_i)$, $i=1,\dots,n$ is as small as possible. Natural functions from $\F$ are $k(\cdot,x)$, $x \in \X$, and thus the goal is to minimize, for such $x \in \X$,
$$
\inf_{\alpha \in \rb^n} \Big\|
\sum_{i=1}^n \alpha_i k(\cdot,x_i) - k(\cdot,x)
\Big\|_\F^2
$$
In the usual sampling approach, several points are considered for testing the projection error, and it is thus natural to consider the criterion averaged through the measure $d \rho$, that is:
$$
\int_\X \inf_{\alpha \in \rb^n} \Big\|
\sum_{i=1}^n \alpha_i k(\cdot,x_i) - k(\cdot,x)
\Big\|_\F^2 d\rho(x).
$$
In fact, when $d\rho$ is supported on a finite set, this formulation is equivalent to minimizing the nuclear norm between the kernel matrix and its low-rank approximation. There are thus several differences and similarities between recent work on column sampling~\citep{bac2012sharp,alaoui2014fast} and the present paper on quadrature rules and random features:
\begin{list}{\labelitemi}{\leftmargin=1.7em}
   \addtolength{\itemsep}{-.215\baselineskip}

\item[--] \textbf{Different error measures}: The column sampling approach corresponds to a function $g$ in \eq{quadrkhs} which is a Dirac function at the point $x$, and is thus not in $L_2(d\rho)$. Thus the two frameworks are not equivalent.

\item[--] \textbf{Approximation vs.~prediction}: The works by~\citet{bac2012sharp,alaoui2014fast} aim at understanding when column sampling leads to no loss in predictive performance within a supervised learning framework, while the present paper looks at approximation properties, mostly regardless of any supervised learning problem, except in \mysec{consequences} for random features (but not for quadrature).

\item[--] \textbf{Lower bounds}: In \mysec{lower}, we provide explicit lower bounds of approximations, which are not available for column sampling.

\item[--] \textbf{Similar sampling issues}: In the two frameworks,   points $x_1,\dots,x_n \in \X$ are sampled i.i.d.~with a certain distribution $q$, and the best choice depends on the appropriate notion of leverage scores~\citep{fot_mahoney}, while the standard uniform distribution leads to an inferior approximation result. Moreover, the proof techniques are similar and based on concentration inequalities for operators, here in Hilbert spaces rather in finite dimensions.

\end{list}

\subsection{Related work on quadrature}

Many methods have been designed for the computation of integrals of a function given   evaluations at certain well-chosen points, in most cases when $g$ is constant equal to one. We review some of these below.

\paragraph{Uni-dimensional integrals.} 
When the underlying set $\X$ is a compact interval of the real line, several methods exists, such as the trapezoidal or Simpson's rules, which are based on interpolation between the sample points, and for which the error decays as $O(1/n^2)$ and $O(1/n^4)$ for functions with uniformly bounded second or fourth derivatives \citep{simpson}. 

Gaussian quadrature is another class of methods for one-dimensional integrals: it is based on a basis of orthogonal polynomials for $L_2( d\rho)$ where $d \rho$ is a probability measure supported in an interval, and their zeros~\citep[Chap.~8]{hildebrand1987introduction}. This leads to quadrature rules which are exact for polynomials of degree $2n-1$ but error bounds for non-polynomials rely on high-order derivatives, although the empirical performance on functions of a Sobolev space in our experiments is as good as optimal quadrature schemes (see \mysec{simu}); depending on the orthogonal polynomials, we get various quadrature rules, such as Gauss-Legendre quadrature for the Lebesgue measure on $[0,1]$.

Quasi Monte-carlo methods employ a sequence of points with low discrepancy with uniform weights \citep{morokoff1994quasi}, leading to approximation errors of $O(1/n)$ for univariate functions with bounded variation, but typically with no adaptation  to smoother functions.

\paragraph{Higher-dimensional integrals.} 
All of the methods above may be generalized for products of intervals $[0,1]^d$, typically with $d$ small. For larger problems, Bayes-Hermite quadrature~\citep{o1991bayes} is essentially equivalent to the quadrature rules we study in this paper.

Some of the quadrature rules are constrained to have positive weights with unit sum (so that the positivity properties of integrals are preserved and constants are exactly inegrated). The quadrature rules we present do not satisfy these constraints. If these constraints are required, kernel herding~\citep{chensuper,bach2012equivalence} provides a novel way to select a sequence of points based on the conditional gradient algorithm, but with currently no convergence guarantees improving over $O(1/\sqrt{n})$ for infinite-dimensional spaces.

\paragraph{Theoretical results.} 
The best possible error for a quadrature rule with $n$ points has been well-studied in several settings; see~\citet{novak1988deterministic} for a comprehensive review. For example, for $\X=[0,1]$ and  the space of Sobolev functions, which are RKHSs with eigenvalues of their integral operator decreasing as $m^{-2s}$,
\citet[Prop.~2 and 3, page 38]{novak1988deterministic} shows that the best possible quadrature rule for the uniform distribution and $g=1$ leads to an error rate of~$n^{-s}$, as well as for any squared-integrable function $g$. The proof of these results (both upper and lower bounds) relies on detailed properties of Sobolev spaces. In this paper, we recover these results using only the decay of eigenvalues of the associated integral operator $\Sigma$, thus allowing straightforward extensions to many situations, like Sobolev spaces on manifolds such as hyperspheres~\citep{hesse2006lower}, where we also recover existing results (up to logarithmic terms).

Moreover, \citet[page 17]{novak1988deterministic} shows that adaptive quadrature rules where points are selected sequentially with the knowledge of the function values at previous points cannot improve the worst-case guarantees. Our results do not recover this lower bound result for adaptivity.

Finally, \citet{landberg} consider multiplicative errors in computing integrals  and mainly focuses on different function spaces, such as ones used in clustering functionals. Although sampling quadrature points from a well-chosen density is common in the two approaches, the analysis tools are different. It would be interesting to see if some of these tools can be transferred to our RKHS setting.

\paragraph{From quadrature to function approximation and optimization.} 
The problem of quadrature, uniformly over all functions $g \in L_2(d\rho)$ that define the integral, is in fact equivalent to the full approximation of a function $h$ given values at $n$ points, where the approximation error is characterized in $L_2$-norm. Indeed, given the observations $h(x_i)$, $i=1,\dots,n$, we build $\sum_{i=1}^n \alpha_i h(x_i)$ as an approximation of $\int_{\X} g(x) h(x) d \rho(x)$. It turns out that the coefficients $\alpha_i$ are linear in $g$, that is, there exists $a_i \in L_2(d\rho)$ such that $\alpha_i  = \langle a_i, g \rangle_{L_2(d\rho)}$. This implies that $\sum_{i=1}^n h(x_i) \langle a_i, g \rangle_{L_2(d\rho)}$ is an approximation of $\langle h, g \rangle_{L_2(d\rho)}$. Thus, the worst case error with respect to $g$ in the unit ball of $L_2(d\rho)$ is
$
\big\| \sum_{i=1}^n h(x_i) a_i - h \big\|_{L_2(d\rho)},
$ that is, we have an approximation result of $h$ through observations of its values at certain points.

 \citet{novak1988deterministic} considers the approximation problem in $L_\infty$-norm and shows that for Sobolev spaces, going from $L_2$- to $L_\infty$-norms incurs a loss of performance of $\sqrt{n}$. We recover partially these results in \mysec{extensions} from a more general perspective. When optimizing the points at which the function is evaluated (adaptively or not), the approximation problem is often referred to as experimental design~\citep{cochran1957experimental,chaloner1995bayesian}. 

Finally, a third problem is of interest (and outside of the scope of this paper), namely the problem of finding the minimum of a function given (potentially noisy) function evaluations. For noiseless problems, \citet[page 26] {novak1988deterministic} shows that the approximation and optimization problems have the same worst-case guarantees (with no influence of adaptivity); this optimization problem has also been studied in the bandit setting~\citep{srinivas2012information} and in the framework of ``Bayesian optimization''~\citep[see, e.g.][]{bull2011convergence}.

\section{Theoretical Analysis}
\label{sec:bounds}

In this section, we provide approximation bounds for the random feature problem outlined in \mysec{approxrand} (and thus the quadrature problem in \mysec{quadd}). In \mysec{upperbound}, we provide generic upper bounds, which depend on the eigenvalues of the integral operator $\Sigma$ and present matching lower bounds (up to logarithmic terms) in \mysec{lowerbound}. The upper-bound depends on specific distributions of samples that we discuss in \mysec{optimized}. We then consider consequences  of these results on quadrature (\mysec{quad}) and random  feature expansions (\mysec{consequences}).

\subsection{Upper bound}
\label{sec:upperbound}
\label{sec:upper}

The following proposition (see proof in Appendix~\ref{app:upper}) determines the minimal number of samples required for a given approximation accuracy:
\begin{proposition}[Approximation of the unit ball of $\F$]
\label{prop:upper}
For $\lambda >0$ and a distribution   with positive density $q$ with respect to $d\tau$, we consider
\BEQ
\label{eq:dmax}
d_{\max}(q,\lambda) =  \sup_{v \in \V}  \frac{1}{q(v)}\langle  \varphi(v,\cdot), ( \Sigma  + \lambda \idm)^{-1}   \varphi(v,\cdot)\rangle_{L_2(d\rho)} .
\EEQ Let $v_1,\dots,v_n$ be sampled i.i.d.~from the density $q$, then for any $\delta \in (0,1)$, if 
$$ \displaystyle n \geqslant 5 d_{\max}(q,\lambda) \log \frac{16  d_{\max} (q,\lambda)}{\delta},$$
  with probability greater than $1-\delta$, we have $\frac{1}{n} \sum_{i=1}^n q(v_i)^{-1} \|  \varphi(v_i,\cdot)  \|_{L_2(d\rho)}^2\leqslant \frac{2 \tr \Sigma}{\delta}$ and
$$\sup_{\| f\|_\F \leqslant 1 } \ \ \inf_{\| \beta \|_2^2 \leqslant \frac{4}{n} }
\bigg\| f - \sum_{i=1}^n \beta_i q(v_i)^{-1/2} \varphi(v_i,\cdot) \bigg\|_{L_2(d\rho)}^2 \leqslant 4 \lambda .$$

\end{proposition}
We can interpret the proposition above as follows: given any squared error $4 \lambda > 0$ and a distribution with density $q$, the number $n$ of samples from $q$ needed so that the unit ball of $\F$ is approximated by the ball of radius $2$ of $\hat{\F}$ is, up to logarithmic terms, at most a constant times $d_{\max}(q,\lambda)$, defined in \eq{dmax}. The result above is a statement for a fixed $q$ and $\lambda$ and this number of samples $n$ depends on these.

We could also invert the relationship between $\lambda$ and $n$, that is, answer the following question: given a fixed number $n$ of samples, what is the approximation error $\lambda$? This requires inverting the function $\lambda \mapsto d_{\max}(q,\lambda)$. This will be done in \mysec{optimized} for a specific distribution $q$ where the expression simplifies, together with specific examples from \mysec{examples}.

Finally, note that we also have a bound on $\frac{1}{n} \sum_{i=1}^n q(v_i)^{-1} \|  \varphi(v_i,\cdot)  \|_{L_2(d\rho)}^2$, which shows that our random functions are not too large on average (this constraint will be needed in the lower bound as well in \mysec{lower}).

\paragraph{Sketch of proof.}
The proof technique relies on computing an explicit candidate $\beta \in \rb^n$ obtained from minimizing a regularized least-squares formulation
$$
\inf_{ \beta \in \rb^n} 
 \Big\| 
\sum_{i=1}^n \beta_i q(v_i)^{-1/2} \varphi(v_i,\cdot) - f 
\Big\|_{L_2(d\rho)}^2 +  {n\lambda}  \| \beta\|_2^2.
$$
It turns out that the final bound on the squared error is exactly proportional to the regularization parameter $\lambda$.
As shown in Appendix~\ref{app:upper}, this leads to an approximation $\hat{f}$ which is a linear function of $f$, as 
$\hat{f} = (\hat{\Sigma}+\lambda \idm)^{-1} \hat{\Sigma} f$, where $\hat{\Sigma}$ is a properly defined empirical integral operator and $\lambda>0$ is the regularization parameter. Then, Bernstein concentration inequalities   for operators~\citep{minsker2011some} can be used in a way similar to the work of~\citet{bac2012sharp,alaoui2014fast} on column sampling, to provide a bound on all desired quantities.

\paragraph{Result in expectation.} In \mysec{consequences}, we will need a result in expectation. As shown at the end of Appendix~\ref{app:upper}, as soons as, $\lambda \leqslant (\tr \Sigma)/4$ and
$ \displaystyle n \geqslant5 d_{\max}(\lambda) \log \frac{  2 (\tr \Sigma) d_{\max} (\lambda)}{\lambda},$
then 
$$ \E \bigg( \sup_{\| f\|_\F \leqslant 1 } \ \ \inf_{\| \beta \|_2^2 \leqslant \frac{4}{n} }
\bigg\| f - \sum_{i=1}^n \beta_i q(v_i)^{-1/2} \varphi(v_i,\cdot) \bigg\|_{L_2(d\rho)}^2 \bigg) \leqslant 8\lambda
.$$

\subsection{Optimized distribution} 
\label{sec:optimized}
We may now consider a specific distribution that depends on the kernel and on $\lambda$, namely 
\BEQ
\label{eq:qopt}
 \displaystyle q^\ast_\lambda(v) = \frac{ \langle \varphi(v,\cdot), ( \Sigma  + \lambda \idm)^{-1}  \varphi(v,\cdot)\rangle_{L_2(d\rho)}}{ \int_\V 
 \langle  \varphi(v,\cdot), ( \Sigma  + \lambda \idm)^{-1}  \varphi(v,\cdot) \rangle_{L_2(d\rho)} d\tau(v) }
 = \frac{ \langle  \varphi(v,\cdot), ( \Sigma  + \lambda \idm)^{-1}  \varphi(v,\cdot)\rangle_{L_2(d\rho)}}{\tr \Sigma(\Sigma +\lambda \idm)^{-1} },
 \EEQ
for which $d_{\max}(q^\ast_\lambda,\lambda) = d(\lambda) =\tr \Sigma(\Sigma +\lambda \idm)^{-1} $.   With this distribution, we thus need to have 
  $n \geqslant 5 d (\lambda) \log \frac{ 16 d (\lambda) }{\delta}  $ with $d (\lambda)= \tr \Sigma(\Sigma +\lambda \idm)^{-1}$ is the \emph{degrees of freedom}, a traditional quantity in the analysis of least-squares regression~\citep{hastie_GAM,Cap_DeV:2007}, which is always smaller than $d_{\max}(1,\lambda)$ and can be upper-bounded explicitly for many examples, as we now explain. The computation of $d_{\max}(1,\lambda)$ in the operator setting (for which we may use $q=1$), a quantity often referred to as the maximal \emph{leverage score}~\citep{fot_mahoney}, remains an open problem.
  
  The quantity $d(\lambda)$ only depends on the integral operator $\Sigma$, that is, for all possible choices of square roots, i.e., all possible choices of feature expansions, the number of samples that our results guarantee is the same. This being said, some expansions may be more computationally practical than others, and when using the distribution with $q(v)=1$, the bounds will be different.

  \paragraph{Expression in terms of singular value decomposition.}
  Given the singular value decomposition of $\varphi$ in \eq{phisvd}, we have, for any $v \in \V$, 
  $\varphi(v,\cdot) = \sum_{m \geqslant 1} \mu_m^{1/2} f_m(v) e_m$ and thus 
  $$
 q^\ast_\lambda(v) \propto 
  \langle \varphi(v,\cdot), ( \Sigma  + \lambda \idm)^{-1}  \varphi(v,\cdot)\rangle_{L_2(d\rho)}
  = \sum_{m \geqslant 1} \frac{\mu_m}{\mu_m + \lambda} f_m(v)^2,
  $$
  which provides an explicit expression for the density $q_\lambda^\ast$.

 For a given squared error value $\lambda$, the optimized distribution $q^\ast_\lambda$, while leading to the degrees of freedom that will happen to be optimal in terms of approximation, has two main drawbacks:
 
 \begin{list}{\labelitemi}{\leftmargin=1.7em}
   \addtolength{\itemsep}{-.215\baselineskip}
\item[--] \textbf{Dependence on $\lambda$}: this implies that if we want a reduced error (i.e., a smaller $\lambda$), then the samples obtained from a higher $\lambda$, may not be reused to provably obtain the desired bound; in other words, the sampling is not \emph{anytime}. For specific examples, e.g., quadrature with periodic kernels on $[0,1]$ with the uniform distribution, then $q=1$ happens to be optimal for all $\lambda$, and thus, we may reuse samples for different values of the error.

\item[--] \textbf{Hard to compute in practice}: the optimal distribution depends on a \emph{leverage score} $ \langle \varphi(v,\cdot), ( \Sigma  + \lambda \idm)^{-1}  \varphi(v,\cdot)\rangle_{L_2(d\rho)}$, which may be hard to use for several reasons; first, it requires access to the infinite-dimensional operator~$\Sigma$, which may be difficult; moreover, even if it possible to invert $\Sigma + \lambda \idm$, the set $\V$ might be particularly large and impractical to sample from. At the end of \mysec{upper}, we propose a simple algorithm based on sampling. 

\end{list}

  \paragraph{Eigenvalues and degrees of freedom.}
  In order  to relate more directly to the eigenvalues of $\Sigma$, we notice that we may lower bound the degrees of freedom by a constant times the number $m^\ast(\lambda)$ of eigenvalues greater than $\lambda$:
  $$
  d(\lambda) =  \tr \Sigma ( \Sigma + \lambda \idm)^{-1}
 = \sum_{m \geqslant 1} \frac{ \mu_m}{\mu_m + \lambda }
 \geqslant \sum_{\mu_m \geqslant \lambda } \frac{ \mu_m}{\mu_m + \lambda }
  \geqslant   \frac{1}{2}  {\rm max}( \{ m  , \ \mu_m   \geqslant \lambda \}) = m^\ast(\lambda),
$$
as defined in \mysec{rkhs}.

 Moreover,  we have the upper-bound:
 $$
d(\lambda)   
 = \sum_{\mu_m \geqslant \lambda } \frac{ \mu_m}{\mu_m + \lambda }
+ \sum_{\mu_m < \lambda } \frac{ \mu_m}{\mu_m + \lambda } 
 \leqslant   {\rm max}( \{ m  , \ \mu_m   \geqslant \lambda \})
+ \frac{1}{\lambda} \sum_{\mu_m < \lambda } \mu_m .
$$
We now make the assumption that there exists a $\gamma >0$ independent of $j$ such that
 \BEQ
\label{eq:mus}  \forall j \geqslant 1, \ \ 
\sum_{m = j}^\infty \mu_{m} \leqslant \gamma j\mu_{j}.
\EEQ
This assumption essentially states that the eigenvalues decay sufficiently homogeneously and is satisfied by $\mu_m \propto m^{-2\alpha}$ with $\gamma = ( 2 \alpha-1)^{-1}$,  $\mu_m \propto r^m$ with $\gamma = ( 1-r)^{-1}$ and similar bounds also hold for all examples in \mysec{examples}. It allows us to relate the degrees of freedom directly to eigenvalue decays.

Indeed, this implies that  $ \frac{1}{\lambda} \sum_{\mu_m < \lambda } \mu_m\leqslant \gamma 
 {\rm max}( \{ m  , \ \mu_m   \geqslant \lambda \}) = m^\ast(\lambda)$ for all $\lambda \leqslant \mu_1$ (the largest eigenvalue) and thus
$$ 
\frac{1}{2} m^\ast(\lambda) \leqslant  d    \leqslant  \big[ 1 + \gamma \big] m^\ast(\lambda)
 .
 $$
  We can now restate the approximation result of Prop.~\ref{prop:upper} from \mysec{upper} with the optimized distribution 
  (see proof in Appendix~\ref{app:upper2}):
  \begin{proposition}[Approximation of the unit ball of $\F$ for optimized distribution]
\label{prop:upper2}
For $\lambda >0$ and the distribution with density   $q^\ast_\lambda$ defined in \eq{qopt}  with respect to $d\tau$, with degrees of freedom $d(\lambda)$.
Let $v_1,\dots,v_n$ be sampled i.i.d.~from the density $q$, defining the kernel (and its associated RKHS $\hat{\F}$) $\hat{k}(x,y) = \frac{1}{n} \sum_{i=1}^n \frac{1}{q(v_i)}
  \varphi({v_i},x) \varphi({v_i},y)$. Then, for any $\delta \in (0,1)$, with probability $1-\delta$, we have:
  $$ \sup_{\| f\|_\F \leqslant 1 } \ \ \inf_{\| \hat{f} \|_{\hat{\F}} \leqslant 2  }
\big\| f - \hat{f}\big\|_{L_2(d\rho)}^2  \leqslant 4 \lambda,$$
under any of the following conditions:
\begin{list}{\labelitemi}{\leftmargin=1.7em}
   \addtolength{\itemsep}{-.215\baselineskip}
   
   \vspace*{-.15cm} 
   
\item[(a)] if $ \displaystyle n \geqslant 5\  d(\lambda) \log \big[ 16  d(\lambda) / \delta \big]$,
\item[(b)] if \eq{mus} is satisfied, and, by choosing $m \leqslant \frac{ n}{5 (1+ \gamma) \log \frac{16 n}{5 \delta}}$, and $\lambda = \mu_m$.
\end{list}
\end{proposition}
 The statement (a) above, is a simple corollary of Prop.~\ref{prop:upper}, and goes from level of error $\lambda$ to minimum number $n$ of samples. The statement (b) goes in the other direction, that is, from the number of samples $n$ to the achieved approximation error. It depends on the eigenvalues $\mu_m$ of the integral operator taken at $m = O( n / \log(n))$.
 For example, for polynomial decays of eigenvalues of the form $\mu_m = O(m^{-2s})$, we get (non squared) errors proportional to $(\log n)^s n^{-s}$ for $n$ samples, while for geometric decays, we get geometric errors as a function of the number $n$ of samples.

Note however that for the statement (b) to hold, we need to sample the points $v_1,\dots,v_n$ from   the distribution $q_{\mu_m}^\ast$, that is, for different numbers of samples $n$, the distribution is unfortunately different (except in special cases). It would be interesting to study the  properties of independent but \emph{not identically distributed} samples $v_1,\dots,v_n$ and the possibility of achieving the same rate adaptively.

\paragraph{Corollary for Sobolev spaces.} For the sake  of concreteness, we consider the special case of $\X = \rb^d$ and translation-invariant kernels. We assume that the distribution $d\rho$ is sub-Gaussian. Then for Sobolev spaces of order $s$, the eigenvalue decay is proportional to $m^{-2s/d}$. Thus, if we can sample from the optimized distribution, after $n$ random features, we obtain   an approximation of the unit ball of $\F$ with error $n^{-s/d}$, independently of the chosen expansion, the spatial one used for quadrature or the spectral one used in random Fourier features. For kernels in $\rb^d$, these distributions are not readily computed in closed form and need to computed through a dedicated algorithm such as the one we present below.

The same approximation results holds for translation-invariant kernels on $[0,1]^d$; but when $d\rho$ is the uniform distribution, as shown in \mysec{quad}, the optimized distribution for the quadrature case is still the uniform distribution, for all values of $\lambda$, and can thus be computed.

 \paragraph{Algorithm to estimate the optimized distribution.}
We now consider a simple algorithm for estimating the optimized distribution $q_\lambda^\ast$. It is based on using a large number $N$ of points $v_1,\dots,v_N$ from $d\tau$, and replacing $d\tau$ by a potentially weighted empirical distribution $d\hat{\tau}$ associated with these $N$ points. Therefore, we may use any set of points and weights,  which leads to a distribution close to $d\tau$. In full generality, only random samples from $d\tau$ are readily available (with weights $1/N$), but for special cases, such as $\V = [0,1]$ or $\V = \mathbb{N}^\ast$, we may use deterministic representations. See examples in \mysec{simu}.

We thus assume that we have $N$ pairs $(v_i,\eta_i) \in \V \times \rb_+$, $i=1,\dots,N$, such that $\sum_{i=1}^n \eta_i = 1$. Since $d\hat\tau$ has a finite support with at most $N$ elements, we may identify $L_2(d\hat\tau)$
and $\rb^N$ (with its canonical dot-product), and the operator $T$ goes now from $\rb^N$ to $L_2(d\rho)$, with $Tg =   \sum_{i=1}^N \eta_i^{1/2} g_i \varphi(v_i,\cdot) \in L_2(d\rho)$, with $T \delta_i = \eta_i^{1/2} \varphi(v_i,\cdot) \in L_2(d\rho)$, for $\delta_i$ the $i$-th element of the canonical basis of $\rb^N$. Then, we have:
\BEAS
 \langle \varphi(v_i,\cdot), ( \Sigma  + \lambda \idm)^{-1}  \varphi(v_i,\cdot)\rangle_{L_2(d\rho)}
 & = &  \eta_i^{-1}   \langle T \delta_i , ( T T^\ast  + \lambda \idm)^{-1} T \delta_i\rangle_{L_2(d\rho)} \\
& = &  \eta_i^{-1}  \langle T \delta_i ,  T ( T^\ast T  + \lambda \idm)^{-1} \delta_i\rangle_{L_2(d\rho)} 
\\
&=& \eta_i^{-1}  \big(
T^\ast T( T^\ast T  + \lambda \idm)^{-1}
\big)_{ii}.
\EEAS
This implies that the density of the optimized distribution with respect to the uniform measure on $\{v_1,\dots,v_N\}$ is proportional to $\big(
T^\ast T( T^\ast T  + \lambda \idm)^{-1}
\big)_{ii}$. We can then sample any number $n$ of points from resampling from $\{v_1,\dots,v_N\}$ from the density above. The computational complexity is $O(N^3)$.  A detailed analysis of the approximation properties of this algorithm is outside the scope of this paper.

We have $(T^\ast T)_{ij} =  {\eta_i^{1/2} \eta_j^{1/2}}\int_\X \varphi(v_i,x) \varphi(v_j,x) d \rho(x)$. In some cases, it can be computed in closed form---such as for quadrature where this is equal to ${\eta_i^{1/2} \eta_j^{1/2}} k(v_i,v_j)$. In some others, it requires i.i.d.~samples $x_1,\dots,x_M$ from $d \rho$, and the estimate:
$ { {\eta_i^{1/2} \eta_j^{1/2}}}{M^{-1}} \sum_{k=1}^M  \varphi(v_i,x_k) \varphi(v_j,x_k)$.

\subsection{Lower bound}
\label{sec:lowerbound}
\label{sec:lower}

In this section, we aim at providing lower-bounds on the number of samples required for a given accuracy. We have the following result (see proof in Appendix~\ref{app:lower}):
\begin{proposition}[Lower approximation bound]
\label{prop:lower}
For $\delta \in (0,1)$, if we have a family $\psi_1,\dots,\psi_n \in L_2(d \rho)$ such that 
$$
\frac{1}{n} \sum_{i=1}^n \| \psi_i\|^2_{L_2(d\rho)} \leqslant 2 \tr \Sigma / \delta,
\ \ \mbox{ and } \ \ \sup_{\| f\|_\F \leqslant 1 } \ \inf_{\| \beta \|_2^2 \leqslant \frac{4}{n} }
\bigg\| f - \sum_{i=1}^n \beta_i \psi_i\bigg\|_{L_2(d\rho)}^2 \leqslant 4 \lambda ,$$
then $\displaystyle n \geqslant \frac{{\rm max}  \{ m  , \ \mu_m   \geqslant 144\lambda \}
 }{ 
4 \log \frac{ 10 \tr \Sigma}{ \lambda \delta }}$.
\end{proposition}
We can make the following observations:

\vspace*{-.15cm}

\begin{list}{\labelitemi}{\leftmargin=1.7em}
   \addtolength{\itemsep}{-.215\baselineskip}

\item[--]
The proof technique not surprisingly borrows tools from minimax estimation over ellipsoids, namely the Varshamov-Gilbert's lemma. 

\item[--] We   obtain matching upper and lower bounds up to logarithmic terms, using only the decay of eigenvalues $(\mu_m)_{m \geqslant 1}$ of the integral operator $\Sigma$ (of course, if sampling from the optimized distribution $q^\ast_\lambda$ is possible). Indeed in that case, as shown in Prop.~\ref{prop:upper2}, we have shown that we need at most 
  $ 10\  d(\lambda) \log \big[ 2 d(\lambda) \big]$, where $d(\lambda)$ is the degrees of freedom, which is upper and lower bounded by a constant times $m^\ast(\lambda) = {\rm max}  \{ m  , \ \mu_m   \geqslant  \lambda \}$.

\item[--] In order to obtain such a bound, we need to constrain both $\|\beta\|_2$ and the norms of the vectors~$\psi_i$, which correspond to bounded   features for the random feature  interpretation and tolerance to noise for the quadrature interpretation. We choose our scaling to match the constraints we have in  Prop.~\ref{prop:upper},  for which   the parameter $\delta$ ends up entering the lower bound logarithmically.

 \end{list}

\subsection{Quadrature}
\label{sec:quad}

We may specialize the results above to the quadrature case, namely give a formulation where the features $\varphi$ do not appear (or equivalently using $\psi$ defined in \mysec{reformulation}). This is a special case where $\V = \X$ and $\varphi = \psi$. In terms of operators $T$ in \mysec{kexp}, this corresponds to $T = \Sigma^{1/2}$.

\paragraph{Optimized distribution.} Following \mysec{upper}, we have an expression for the optimized distribution, both in terms of operators, as follows,
$$
q_\lambda^\ast(x) \propto \langle \psi(x,\cdot), ( \Sigma + \lambda \idm)^{-1} \psi(x,\cdot) \rangle_{L_2(d\rho)}
= \langle \Sigma^{-1/2}k(x,\cdot), ( \Sigma + \lambda \idm)^{-1} \Sigma^{-1/2}k(x,\cdot) \rangle_{L_2(d\rho)},
$$
and in terms of eigenvalues and eigenvectors of $k$, that is,
\BEQ
\label{eq:sobq}
q(x) \propto \langle k(\cdot,x), \Sigma^{-1/2} (\Sigma+\lambda \idm)^{-1} \Sigma^{-1/2}  k(\cdot,x) \rangle_{L_2(d\rho)} =\sum_{m \geqslant 1} \frac{\mu_m}{\mu_m + \lambda} e_m(x)^2.
\EEQ
While this is uniform in some special cases (uniform distribution on $[0,1]$ and Sobolev kernels, as shown below), this is typically hard to compute and sample from. An algorithm for approximating it was presented at the end of \mysec{upper}.

A weakness of our result is that in general our optimized distribution $  q^\ast_\lambda(x)$ depends on $\lambda$ and thus on the number of samples. In some cases with symmetries (i.e., uniform distribution on $[0,1]$ or the hypersphere), $q^\ast_\lambda$ happens to be constant for all $\lambda$. Note also that we have observed empirically that in some cases,  $q^\ast_\lambda$ converges to a certain distribution when $\lambda $ tends to zero (see an example in \mysec{simu}).

\paragraph{Sobolev spaces.} 
For Sobolev spaces with order $s$ in $[0,1]^d$  or $\rb^d$ (for which we assume $d<2s$), the decay of eigenvalues is of the form $m^{-2s/d}$ and thus the error after $n$ samples is $n^{-s/d}$ (up to logarithmic terms), which recovers the upper and lower bounds of \citet[pages~37 and~38]{novak1988deterministic} (also up to logarithmic terms).  

For the special case of Sobolev spaces on $[0,1]^d$ with $d\rho$ the uniform distribution, the optimized distribution in \eq{sobq} is  also the uniform distribution. Indeed, the eigenfunctions of the integral operator $\Sigma$ are $d$-th order tensor products of the uni-dimensional Fourier basis (the constant and all pairs of sine/cosine at a given frequency), with the \emph{same eigenvalue} for the $2^d$ possibilities of sines/cosines for a given multi-dimensional frequency $(m_1,\dots,m_d)$. Therefore, when summing all squared values of the eigenfunctions corresponding to $(m_1,\dots,m_d)$, we end up with the sum $\sum_{a \in \{0,1\}^d} \prod_{i=1}^d\cos^{2a_i}( 2\pi m_i x_i) 
\sin^{2(1-a_i)}( 2\pi m_i x_i)$, which ends up being constant equal to one (and thus independent of $x$) because 
$\cos^{2a_i}( 2\pi m_i x_i) + \sin^{2a_i}( 2\pi m_i x_i) = 1$.

Finally, we may consider Sobolev spaces on the hypersphere, with the kernels presented in \mysec{examples}. As shown by~\citet[Appendix D.3]{relu}, the kernel $k(x,y) = \E (v^\top y)_+^s (v^\top y)_+^s$ for $v$ uniform on the hypersphere, leads to a Sobolev space of order $t = s + \frac{d+1}{2}$, while the decay of eigenvalue of the integral operator was shown to be 
$m^{-1-1/d - 2s/d}$ in \mysec{examples}. It is thus equal to $m^{-2t/d}$, and we recover the result from~\citet{hesse2006lower}.

\paragraph{Quadrature rule.} We assume that points $x_1,\dots,x_n$ are sampled from the distribution with density $q$ with respect to $d\rho$. The quadrature rule for a function $h \in \F$ is $ \sum_{i=1}^n \frac{\beta_i h(x_i)}{q(x_i)^{1/2}}$. To compute $\beta$, we need to minimize with respect to $\beta$ the error:
$$\bigg\| 
\sum_{i=1}^n \frac{\beta_i }{q(x_i)^{1/2}} k(\cdot,x_i) - \int_\X k(\cdot,x) g(x) d \rho(x)
\bigg\|_\F^2 + n \lambda \| \beta\|_2^2,
$$
which is the regularized worst case squared error in the estimation of the integral of $h$ over $h \in \F$. The best error is obtained for $\lambda=0$, but our guarantees are valid for $\lambda >0$, with an explicit control over the norm $\| \beta\|_2^2$, which is important for robustness to noise.  

Given the values of $\int_\X k(x_i,x) g(x) d \rho(x) = z_i$, for $i=1,\dots,n$,
which can be computed in closed form for several triplet $(k,g,d\rho)$~\citep[see, e.g.,][]{smola2007hilbert,oates2015variance},
 then the problem above is equivalent to minimizing with respect to $\beta$:
$$
\sum_{i=1}^n \sum_{j=1}^n \frac{\beta_i \beta_j }{q(x_i)^{1/2}q(x_j)^{1/2}} k(x_i,x_j) -
\sum_{i=1}^n \frac{\beta_i   }{q(x_i)^{1/2}} z_i + n \lambda \| \beta\|_2^2,
$$
which leads to a $n\times n$ linear system with running time complexity $O(n^3)$.
Note that when adding points sequentially (in particular for kernels for which the distribution $q_\lambda^\ast$ is independent of $\lambda$, such as  Sobolev spaces on $[0,1]$), one may update the solution so that after $n$ steps, the overall complexity is $O(n^3)$.

\paragraph{Approximation of functions in $\F$.}
With the quadrature weights $\beta$ estimated above and the quadrature rule $ \sum_{i=1}^n \frac{\beta_i h(x_i)}{q(x_i)^{1/2}}$
for the estimation of $\int_{\X} g(x) f(x) d \rho(x)$, we may derive an expression which is explicitly linear in $g$. 
Following the proof of Prop.~\ref{prop:upper} in Appendix~\ref{app:upper}, we have, when specialized to the quadrature case:
$$
\hat{\Sigma} =   \frac{1}{n}\sum_{i=1}^n \frac{1}{q(v_i)} \psi(x_i,\cdot)  \otimes_{L_2(d\rho)} \psi(x_i,\cdot) = \Sigma^{-1/2} \bigg(
 \frac{1}{n}\sum_{i=1}^n \frac{1}{q(v_i)} k(x_i,\cdot)  \otimes_{L_2(d\rho)} k(x_i,\cdot) \bigg) \Sigma^{-1/2},
$$
Moreover, we have $\beta_i = \frac{1}{n q(x_i)^{1/2}} \langle k(\cdot,x_i), \Sigma^{-1/2} (\hat{\Sigma} + \lambda \idm)^{-1}   \Sigma^{1/2} g\rangle_{L_2(d\rho)}$  from \eq{beta} in Appendix~\ref{app:upper}, and the quadrature rule becomes:
\BEAS
 \sum_{i=1}^n \frac{\beta_i h(x_i)}{q(x_i)^{1/2}} \!\!\!
& = \!\!\! & \sum_{i=1}^n \frac{\beta_i}{q(x_i)^{1/2}} \langle h, \Sigma^{-1} k(\cdot,x_i)\rangle_{L_2(d\rho)} \\
& = \!\!\!& \bigg\langle
h, 
\frac{1}{n} \sum_{i=1}^n \Sigma^{-1} \frac{1}{q(x_i)} 
\big[ k(x_i,\cdot)  \otimes_{L_2(d\rho)} k(x_i,\cdot)  \big] \Sigma^{-1/2} (\hat{\Sigma} + \lambda \idm)^{-1}  \Sigma^{1/2} g
\bigg\rangle_{L_2(d\rho)} \\
& = \!\!\!& \big\langle
h,   \Sigma^{-1/2} \hat{\Sigma}   (\hat{\Sigma} + \lambda \idm)^{-1} \Sigma^{1/2} g
\big\rangle_{L_2(d\rho)} 
=  \big\langle
g,   \Sigma^{ 1/2} \hat{\Sigma}   (\hat{\Sigma} + \lambda \idm)^{-1}   \Sigma^{-1/2} h
\big\rangle_{L_2(d\rho)} ,
\EEAS
which can be put in the form $\langle \hat{h}, g \rangle_{L_2(d\rho)} $ with the approximation $\hat{h}=  \Sigma^{ 1/2} \hat{\Sigma}   (\hat{\Sigma} + \lambda \idm)^{-1}   \Sigma^{-1/2} h$ of the function $h \in \F$. Having a bound for all functions $g$ such that $\|g\|_{L_2(d\rho)}\leqslant 1$ is equivalent to having a bound on $\|h-\hat{h}\|_{L_2(d\rho)}$.
In \mysec{extensions}, we consider extensions, where we consider other norms than the $L_2$-norm for characterizing the approximation error $\hat{h} - h$. Moreover, we consider cases where $h$ belongs to a strict subspace of $\F$ (with improved results).

\subsection{Learning with random features}
\label{sec:consequences}

We consider supervised learning with $m$ i.i.d.~samples from a distribution on  inputs/outputs $(x,y)$, and a uniformly $G$-Lipschitz-continuous loss function $\ell(y,\cdot)$, which includes logistic regression and the support vector machine. We consider the empirical risk $\hat{L}(f) = \frac{1}{m} \sum_{i=1}^m \ell(y_i,f(x_i))$ and the expected risk $L(f) = \E \ell(y,f(x))$, with $x$ having the marginal distribution $d\rho$ that we consider in earlier sections. We assume that $\E k(x,x) = \tr \Sigma =  R^2$. We have the usual generalization bound for the minimizer $\hat{f}$ of $\hat{L}(f)$ with respect to $\| f\|_\F \leqslant F$, based on Rademacher complexity \citep[see, e.g.,][]{shaibook}:
\BEQ
\label{eq:boundF}
\E \big[ L(\hat{f}) \big] \leqslant    \inf_{\| f\|_\F \leqslant F} L(f) + 2 \E \Big[ \sup_{\| f\|_\F \leqslant F} | L(f) - \hat{L}(f) |\Big]
  \leqslant    \inf_{\| f\|_\F \leqslant F} L(f) +   \frac{4 FGR}{\sqrt{m}}.
\EEQ

We now consider learning  by sampling $n$ features from the optimized distribution from \mysec{optimized}, leading to a function parameterized by $\beta \in \rb^n$,  that is $\hat{g}_\beta = \sum_{i=1}^n \beta_i q(v_i)^{-1/2}\varphi(v_i,\cdot) \in L_2(d\rho)$. Applying results from \mysec{upper}, we assume that
$\lambda \leqslant R^2/4 $ and $n \geqslant 5 d (\lambda) \log \frac{2  (\tr \Sigma) d  (\lambda)}{\lambda}$, where $d(\lambda)$ is equal to the degrees of freedom associated with the kernel $k$ and distribution $d\rho$. Thus, the expected squared error for approximating the unit-ball of $\F$ by the ball of radius $2$ of the approximation $\hat{\F}$ obtained from the approximated kernel is less than $8\lambda$.

If we consider the estimator $\hat\beta$ obtained by minimizing the empirical risk of $\hat{g}_\beta$  subject to $\| \beta \|_2 \leqslant 2 F / \sqrt{n}$. We have the following decomposition of the error for any $\gamma \in \rb^n$ such that 
$\| \gamma \|_2 \leqslant 2 F / \sqrt{n}$ and $f \in \F$ such that $\| f\|_\F \leqslant F$:
\BEAS
L(\hat{g}_{\hat{\beta}})
& = & L(\hat{g}_{\hat{\beta}}) - \hat{L}(\hat{g}_{\hat{\beta}})
+ \hat{L}(\hat{g}_{\hat{\beta}}) - \hat{L}(\hat{g}_{\gamma})
+\hat{L}(\hat{g}_{\gamma}) -  {L}(\hat{g}_{\gamma})
+ {L}(\hat{g}_{\gamma}) - L(f) + L(f) \\
& \leqslant & 2  \Big[ \sup_{\| \beta'\|_\F \leqslant 2F / \sqrt{n}} | L(\hat{g}_{ {\beta}'})  - L(\hat{g}_{ {\beta}'}) |\Big]
+ \big[ {L}(\hat{g}_{\gamma}) - L(f) \big]  + L(f) \\
& \leqslant & 2  \Big[ \sup_{\| \beta'\|_\F \leqslant 2F / \sqrt{n}} | L(\hat{g}_{ {\beta}'})  - L(\hat{g}_{ {\beta}'}) |\Big]
+ \sup_{\|f'\|_\F \leqslant  F} \inf_{ \| \gamma\|_2 \leqslant 2 F / \sqrt{n} } \big[ {L}(\hat{g}_{\gamma}) - L(f') \big]  + \inf_{\| f\|_\F \leqslant F} L(f) .
\EEAS
We now take expectation with respect to the data and the random features.
 Following standard results for Rademacher complexities of $\ell_2$-balls~\citep[][Lemma 22]{bartlett2003rademacher}, the first term is less than
$$ \frac{4FG}{m \sqrt{n}} \E \big( \sum_{i=1}^m \sum_{j=1}^n \frac{ \varphi(v_i,x_j)^2}{q(v_i)} \big)^{1/2}
\leqslant  \frac{4FG}{m \sqrt{n}} (n m \tr \Sigma)^{1/2} =   \frac{4FGR}{  \sqrt{m}} .$$
Because of the $G$-Lipschitz-continuity of the loss,
we have $ {L}(\hat{g}_{\gamma}) - L(f')  \leqslant  {G} \| \hat{g}_{\gamma}) - f'\|_{L_2(d\rho)}$, and thus the second term is less than $\sqrt{ 8 \lambda } G F \leqslant 3 GF \sqrt{\lambda}$. Overall, we obtain
$$
\E \big[ L(\hat{g}_{\hat{\beta}}) \big] \leqslant   \inf_{\| f\|_\F \leqslant F} \!  L(f) + 3  GF \sqrt{ \lambda} +
 \frac{4FGR}{  \sqrt{m}}.$$
If we consider $\lambda = R^2 / m$ in order to lose only a constant factor compared to \eq{boundF}, we have the constraint 
  $n \geqslant 5 d (R^2/m) \log \big[ {2 m d  (R^2/m)}\big]$.
    
  We may now look at several situations. In the worst case, where the decay of eigenvalue is not fast, i.e., very close to $1/i$, then we may only use the bound $d(\lambda) = \tr \Sigma (\Sigma + \lambda \idm)^{-1} \leqslant \lambda^{-1} \tr \Sigma= R^2  / \lambda$, and thus a sufficient condition
 $ n \geqslant  10  m \log 2 m$, and we obtain the same result as \citet{rahimi2009weighted}. 
 
 However, when we have eigenvalue decays as $R^2 i^{-2s}$, we get (up to  constants), following the same computation as \mysec{optimized}, $d(\lambda) \leqslant (R^2/\lambda )^{1/(2s)}$, and thus $n \geqslant  m^{1/(2s)} \log m$, which is a significant improvement (regardless of the value of $F$). Moreover, if the decay is geometric as~$r^i$, then we get $d(\lambda) \leqslant \log (R^2/\lambda ) $, and thus $n \geqslant  (\log m)^2 $, which is   even more significant.

 \section{Quadrature-related Extensions}
 \label{sec:extensions}

  In \mysec{quad}, we have built an approximation
  $\hat{h}=  \Sigma^{ 1/2} \hat{\Sigma}   (\hat{\Sigma} + \lambda \idm)^{-1}   \Sigma^{-1/2} h$ of a function $h \in \F$, which is based on $n$ function evaluations $h(x_1),\dots,h(x_n)$. We have presented in \mysec{quad} a convergence rate for the $L_2$-norm $\| \hat{h} - h\|_{L_2(d\rho)}$ for functions $h$ with less than unit $\F$-norm $\| h \|_\F \leqslant 1$. Up to logarithmic terms, if using the optimal distribution for sampling $x_1,\dots,x_n$, then we get a squared error of $\mu_n$ where $\mu_n$ is the $n$-th largest eigenvalue of the integral operator $\Sigma$.

 \paragraph{Robustness to noise.} 
We have seen that if the noise in the function evaluations $h(x_i)$ has a variance less than $q(x_i) \tau^2$, then  the error $\| h - \hat{h}\|_{L_2(d\rho)}^2$ has an additional term $\tau^2 \| \beta\|_2^2 \leqslant \frac{4 \tau^2}{n}$. Hence, the amount of noise has to be less than $n \mu_n$ in order to incur no loss in performance (a bound which decreases with $n$).
 
 \paragraph{Adaptivity to smoother functions.} 
 
We assume that the function $h$ happens to be smoother than what is sufficient to be an element of the RKHS $\F$, that is, if $\|\Sigma^{-s} h\|_{L_2(d\rho)} \leqslant 1$, where $s \geqslant 1/2$. The case $s = 1/2$ corresponds to being in the RKHS.
In the proof of Prop.~\ref{prop:upper} in Appendix~\ref{app:upper}, we have seen that with high-probability we have: 
\BEQ
 \label{eq:S} (\hat{\Sigma} + \lambda \idm)^{-1} \preccurlyeq 4  ( {\Sigma} + \lambda \idm)^{-1}
 .\EEQ
  We now see that we can bound the error 
$ \| \hat{h} - h\|_{L_2(d\rho)}$ as follows:
 \BEAS
 \| \hat{h} - h\|_{L_2(d\rho)}
 & = &  \| \Sigma^{ 1/2} \hat{\Sigma}   (\hat{\Sigma} + \lambda \idm)^{-1}   \Sigma^{-1/2} h - h\|_{L_2(d\rho)}  \\
 & = & \lambda \big\|  \Sigma^{ 1/2}     (\hat{\Sigma} + \lambda \idm)^{-1}  \Sigma^{-1/2+s}  \Sigma^{-s} h \big\|_{L_2(d\rho)} \\
& \leqslant &\lambda  \big\|  \Sigma^{ 1/2}     (\hat{\Sigma} + \lambda \idm)^{-1/2} \big\|_{\rm op}
\big\|    (\hat{\Sigma} + \lambda \idm)^{-1/2}  \Sigma^{-1/2+s} \big\|_{\rm op}
\| \Sigma^{-s} h  \|_{L_2(d\rho)}.
\EEAS
We may now bound each term. The first one $\big\|  \Sigma^{ 1/2}     (\hat{\Sigma} + \lambda \idm)^{-1/2} \big\|_{\rm op}$ is less than 2, because of \eq{S}. The second one $\big\|    (\hat{\Sigma} + \lambda \idm)^{-1/2}  \Sigma^{-1/2+s} \big\|_{\rm op}
$ is equal to $\big\|   (\hat{\Sigma} + \lambda \idm)^{s-1}   (\hat{\Sigma} + \lambda \idm)^{ 1/2 - s }  \Sigma^{-1/2+s} \big\|_{\rm op}
$, and thus less than $\big\|   (\hat{\Sigma} + \lambda \idm)^{s-1} \|_{\rm op} \cdot \big\|    (\hat{\Sigma} + \lambda \idm)^{1/2-s}  \Sigma^{-1/2+s} \big\|_{\rm op} \leqslant 2 \lambda^{s-1}$. Overall we obtain
$$
\| \hat{h} - h\|_{L_2(d\rho)}
 \leqslant   4 \lambda^{s }.
 $$
The norm $h \mapsto \|\Sigma^{-s} h\|_{L_2(d\rho)}$ is an RKHS norm with kernel $\sum_{m \geqslant 0} \mu_m^{2s} e_m(x) e_m(y)$, with corresponding eigenvalues equal
 to $(\mu_m)^{2s}$. From Prop.~\ref{prop:upper2} and \ref{prop:lower}, the optimal number of quadrature points to reach a squared error less than $\varepsilon$ is proportional to the number
 $ {\rm max}( \{ m  , \ \mu_m^{2s}   \geqslant \varepsilon \}) $, while using the quadrature points from $s=1/2$, leads to a number  $  {\rm max}( \{ m  , \ \mu_m    \geqslant \varepsilon^{1/(2s)} \})$, which is equal.
 Thus if the RKHS used to compute the quadrature weights is a bit too large (but not too large, see experiments in \mysec{simu}), then we still get the optimal rate. Note that this robustness is only shown for the regularized estimation of the quadrature coefficients (in our simulations, the non-regularized ones also exhibit the same behavior).

 \paragraph{Approximation with stronger norms.} 
 We may consider characterizing the difference $\hat{h}-h$ with different norms than $\|\cdot\|_{L_2(d\rho)}$, in particular norms $\| \Sigma^{-r} ( \hat{h} - h)\|_{L_2(d\rho)}$, with $ r \in [0,1/2]$. For $r=0$, this is our results in $L_2$-norm, while for $r=1/2$, this is the RKHS norms.
 We have, using the same manipulations than above:
 \BEAS
\| \Sigma^{-r} ( \hat{h} - h)\|_{L_2(d\rho)}
 & = & \lambda \big\|  \Sigma^{ 1/2-r}     (\hat{\Sigma} + \lambda \idm)^{-1}  \Sigma^{-1/2}   h \big\|_{L_2(d\rho)} \\
& \leqslant &\lambda^{1/2-r}  \big\|  \Sigma^{ 1/2-r}     (\hat{\Sigma} + \lambda \idm)^{r-1/2} \big\|_{\rm op} 
\| \Sigma^{-1/2} h  \|_{L_2(d\rho)}  
 \leqslant     2 \lambda^{1/2-r }    .
 \EEAS
When $r=1/2$, we get a result in the RKHS norm, but with no decay to zero; the RKHS norm $\| \cdot \|_\F$ would allow a control in $L_\infty$-norm, but as noticed by \citet{steinwart2009optimal,mendelson2010}, such a control may be obtained in practice with $r$ much smaller. For example, when the eigenfunctions $e_m$ are uniformly bounded in $L_\infty$-norm by a constant $C$ (as is the case for periodic kernels in $[0,1]$ with the uniform distribution), then, for any $x \in \X$, we have for $t>1$,
$$
f(x)^2 = \sum_{m=1}^\infty (m+1)^t \langle f, e_m \rangle^2_{L_2(d\rho)}
  e_m(x)^2 (m+1)^{-t} \leqslant
 \sum_{m=0}^\infty (m+1)^t \langle f, e_m \rangle^2_{L_2(d\rho)}
\frac{C^2}{t - 1}.
$$
If for simplicity, we assume that $\mu_m =  (m+1)^{-2s}$ (like for Sobolev spaces), we have
 $   \| \Sigma^{-r} f\|_{L_2(d\rho)}^2 =
  \sum_{m=1}^\infty \mu_m^{-2r} \langle f, e_m \rangle^2_{L_2(d\rho)}
  =
  \sum_{m=1}^\infty (m+1)^t \langle f, e_m \rangle^2_{L_2(d\rho)}$ with $r = t / 4 s$.
 If $\lambda \leqslant O(n^{-2s})$ (as suggested by Prop.~\ref{prop:upper}), then we obtain a squared $L_\infty$-error less than
 $
 \frac{1}{t - 1}
 \lambda^{1-2r} = O\big( \frac{1}{t - 1} n^{-2s ( 1 - t / 2 s) } \big)
=  O\big( \frac{n^t}{t - 1} n^{-2s}   \big)
 $.
 With $t = 1 + \frac{1}{\log n}$, we get
 $ O\big( \frac{n \log n}{n^{-2s}}   \big)$,
 and thus a degradation compared to the  squared $L_2$-loss of $n $ (plus additional logarithmic terms), which corresponds to the (non-improvable) result of \citet[page 36]{novak1988deterministic}.

 \section{Simulations}

\label{sec:simu}
\label{sec:simulation}
\label{sec:simulations}
In this section, we consider simple illustrative quadrature experiments\footnote{Matlab code for all 5 figures may be downloaded from~{\scriptsize\url{http://www.di.ens.fr/~fbach/quadrature.html}}.} with $\X = [0,1]$ and kernels $k(x,y)
= 1 + \sum_{m=1}^\infty \frac{1}{m^{2s}} \cos 2\pi m (x-y)$, with various values of $s$ and distributions $d\rho$ which are Beta random variable with the two  parameters equal to $a=b$, hence symmetric around $1/2$.

\begin{figure}
\begin{center}
\includegraphics[scale=.5]{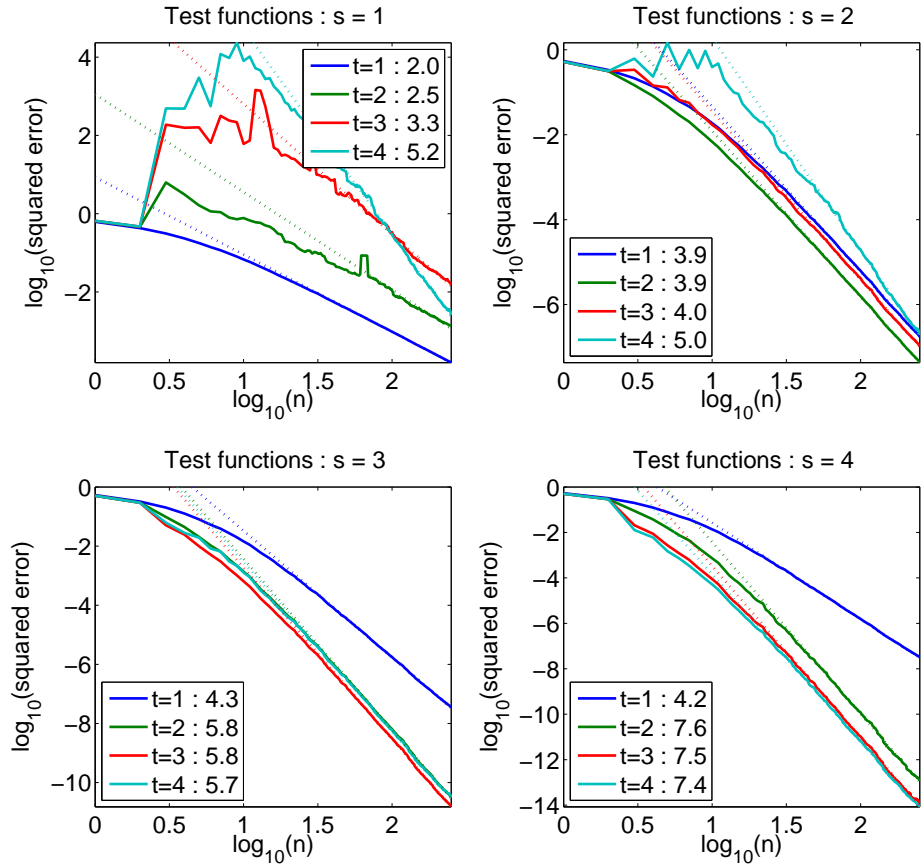}
\end{center}

\vspace*{-.25cm}

\caption{Quadrature for functions in a Sobolev space with parameter $s$ (four possible values) for the uniform distribution on $[0,1]$, with quadrature rules obtained from different Sobolev spaces with parameters $t$ (same four possible values). We compute affine fits in log-log-space (in dotted) to estimate convergence rates of the form $C/n^u$ and report the value of $u$. Best seen in color.}
\label{fig:sobolev}
\end{figure}

\paragraph{Uniform distribution.} 
For $b=1$, we have the uniform distribution on $[0,1]$ for which the cosine/sine basis is orthonormal, and the optimized distribution $q_\lambda^\ast$ is also uniform. Moreover, we have $
\int_0^1 k(x,y) d\rho(x) = 1
$. We report results comparing different Sobolev spaces for testing functions to integrate  (parameterized by $s$)  and learning quadrature weights (parameterized by $t$) in Figure~\ref{fig:sobolev}, where we compute errors averaged over 1000 draws. We did not use regularization to compute quadrature weights $\alpha$. We can make the following observations:
\BIT
\item[--] The exponents in the convergence rates for $s=t$ (matching RKHSs for learning quadrature weights and testing functions) are close to $2s$ as expected.
\item[--] When the functions to integrate are less smooth than the ones used for learning quadrature weights (that is $t > s$), then the quadrature performance does not necessarily decay with the number of samples.
\item[--] On the contrary, when $s > t$, then we have convergence and the rate is potentially worse than the optimal one (attained for $s=t$), and equal when $t \geqslant s/2$, as shown in \mysec{extensions}.
\EIT

\begin{figure}
\begin{center}
\includegraphics[scale=.5]{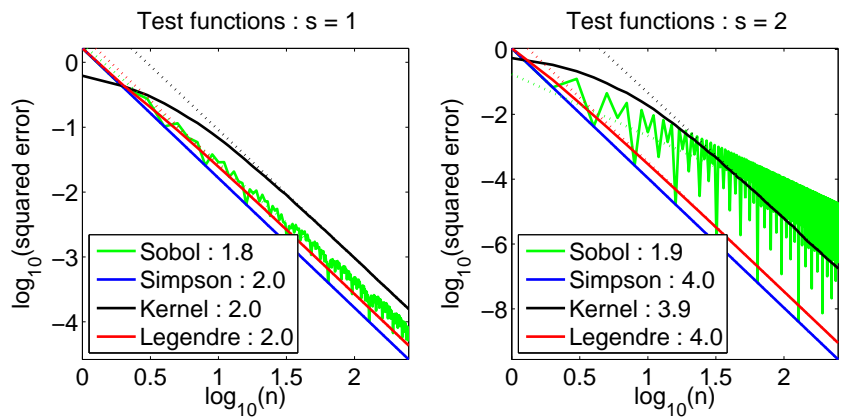}
\end{center}

\vspace*{-.25cm}

\caption{Quadrature for functions in a Sobolev space with parameters $s=1$ (left) and $s=2$ (right), for the uniform distribution on $[0,1]$, with various quadrature rules. We compute affine fits in log-log-space (in dotted) to estimate convergence rates of the form $C/n^u$ and report the value of $u$. Best seen in color.}
\label{fig:comparison}
\end{figure}

In Figure~\ref{fig:comparison}, we compare several quadrature rules on $[0,1]$, namely Simpson's rule with uniformly spread points, Gauss-Legendre quadrature and the Sobol sequence with uniform weights. For $s=1$, as expected, all squared errors decay as $n^{-2}$ with a worse constant for our kernel-based rule, while for $s=2$ (smoother test functions), the Sobol sequence is not adaptive, while all others are adaptive and get convergence rates around $n^{-4}$.

\paragraph{Non-uniform distribution.}  
We consider the case $a=b=1/2$, which is the distribution $d\rho$ with density $  \pi^{-1} x^{-1/2} ( 1 - x)^{-1/2}$ with respect to the Lebesgue measure, and with cumulative distribution function $F(x) = \pi^{-1} \arccos(1-2x)$. We may use an approximation of $d\tau$ with $N$ unweighted points $F^{-1}(k/N) = \big( 1 - \cos \frac{k\pi}{N} \big)/2$, for $k \in \{1,\dots,N\}$ and the algorithms from the end of \mysec{optimized}. We consider the Sobolev kernel with $s=1$.

\begin{figure}
\begin{center}
\includegraphics[scale=.4]{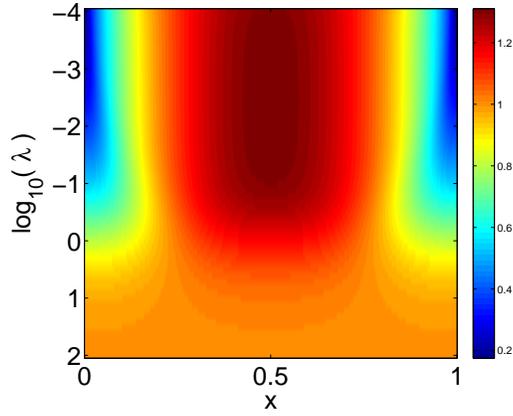}
\end{center}
\caption{Optimal log-densities $q_\lambda^\ast(x)$ (with respect to the input distribution) for several values of $\lambda$, for the expansion used for quadrature.  
Best seen in color.}
\label{fig:beta}

\end{figure}

In \myfig{beta}, we plot all densities $q_\lambda^\ast$ as a function of $\lambda$. When $\lambda$ is large, we unsuprisingly obtain the uniform density, while, more surprisingly, when $\lambda$ tends to zero, the density tends to a density, which happens here to be proportional to $x^{1/4} (1-x)^{1/4}$ (leading to a Beta distribution with parameters $a=b=.25$).

\begin{figure}
\begin{center}
\includegraphics[scale=.4]{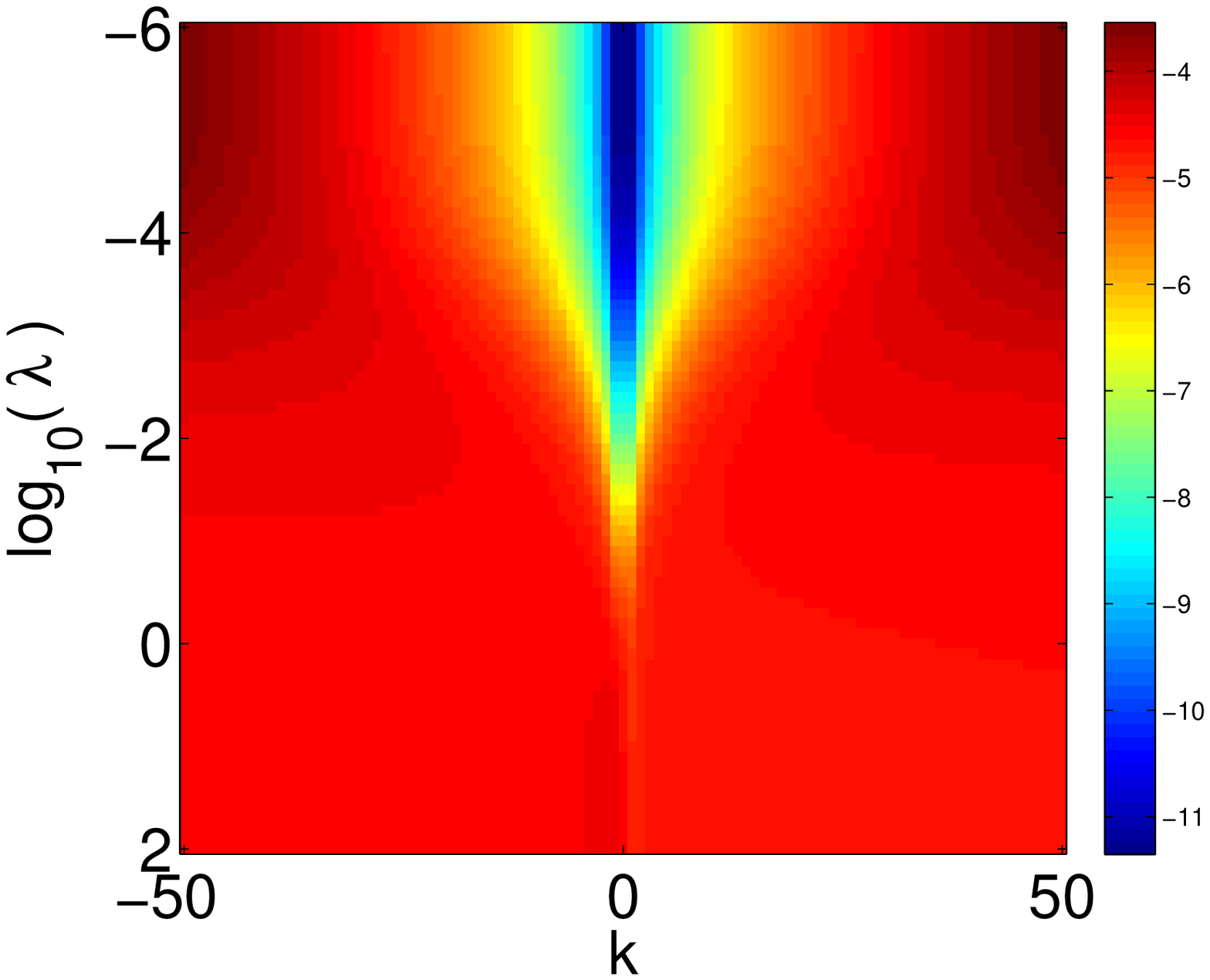} \hspace*{.5cm}
\includegraphics[scale=.4]{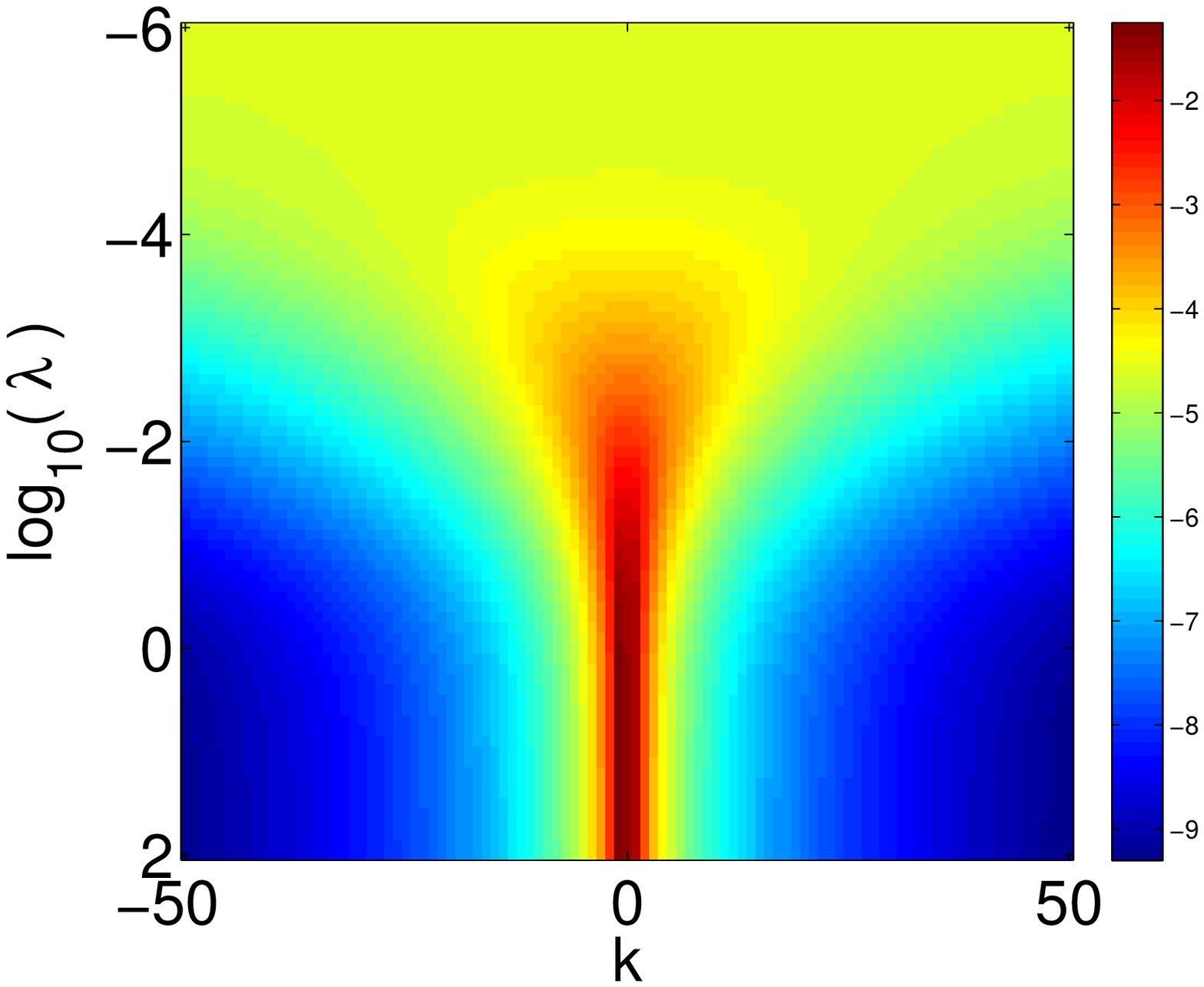}
\end{center}
\caption{Optimal densities $q_\lambda^\ast(k)$ for several values of $\lambda$, for Fourier feature expansions. Left: with respect to the input distribution (which itself has distribution proportional to $1/k^2$ with respect to the counting measure); right: with respect to the counting measure.
Best seen in color.}
\label{fig:betafourier}

\end{figure}

We may also consider the same kernel but with the Fourier expansion on $\mathbb{N}$.
This is done by representing $d\tau \propto \delta_0 + \sum_{k \in \mathbb{Z}^\ast} \frac{1}{k^2} \delta_k $ by truncating to all $|k| \leqslant K$, with $K=50$, which is a weighted representation.
 We plot in \myfig{betafourier} the optimal density over the set of integers, both with respect to the input density (which decays as $1/n^2$) and the counting measure. When $\lambda$ is large, we recover the input density, while when $\lambda$ tends to zero, $q_\lambda^\ast$ tends to be uniform (and thus, does not converge to a finite measure).

\section{Conclusion}

In this paper, we have shown that  kernel-based quadrature rules are a special case of random feature expansions for positive definite kernels, and derived upper and lower bounds on approximations, that match up to logarithmic terms. For quadrature, this leads to widely applicable results while for random features this allows a significantly improved guarantee within a supervised learning framework.

The present work could be extended in a variety of ways, for example towards bandit optimization rather than quadrature~\citep{srinivas2012information}, the use of quasi-random sampling within our framework in the spirit of~\citet{yang2014quasi,oates2015variance}, a similar analysis for kernel herding~\citep{chensuper,bach2012equivalence}, an extension to fast rates for non-parametric least-squares regression~\citep{hsu2014random} but with an improved computational complexity, and a study of the consequences of our improved approximation result for online learning and stochastic approximation, in the spirit of \citet{dai2014scalable,klms}.
 
\subsection*{Acknowledgements}
This work was partially supported
by the MSR-Inria Joint Centre and a grant by the European Research Council (SIERRA project 239993). Comments of the reviewers were greatly appreciated and helped improve the presentation significantly.
The author would like to thank the STVI for the opportunity of writing a single-handed paper.

\bibliography{relu}

\appendix

 \section{Kernels on product spaces}
 \label{app:product}
 In this appendix, we consider sets $\X$ which are products of several simple sets $\X_1,\dots,\X_d$, with known kernels $k_1,\dots,k_d$, each with RKHS $\F_1,\dots,\F_d$. We also assume that we have $d$ measures $d \rho_1,\dots,d \rho_d$, leading to sequences of eigenvalues $(\mu_{j m_j})_{m_j \geqslant 1}$ and eigenfunctions $(e_{jm_j})_{m_j \geqslant 1}$.
 
 Our aim is to define a kernel $k$ on $\X = \X_1 \times \cdots \times \X_d$ with the product measure $d \rho = d\rho_1 \cdots d \rho_d$. For illustration purposes, we consider decays of the form $\mu_m \propto m^{-2s}$ for the $d$ kernels, that will be useful for Sobolev spaces. We also consider the case where $\mu_m \propto \exp( - \rho m)$. For some combinations, eigenvalue decay is the most natural, in others, the number of eigenvalues $m^\ast(\lambda)$ greater than a given  $\lambda>0$ is more natural.

\subsection{Sum of kernels: $k(x,y) = \sum_{j=1}^d k_j(x_j,y_j)$}

 In this situation, the RKHS   for $k$ is isomorphic to 
$\F_1 \times \cdots \times \F_d$, composed of functions  $g$ such that there exists $f_1,\dots,f_d $ in $\F_1,\dots,\F_d$ such that $g(x) = \sum_{j=1}^d f_j(x_j)$, that is we obtain separable functions, which are sometimes used in the context of generalized additive models~\citep{hastie_GAM}.
 The corresponding integral operator  is then block-diagonal with $j$-th block equal to the integral operator for $k_j$ and $d\rho_j$. This implies that
that its eigenvalues are the concatenation of all sequences $(\mu_{j m_j})_{m_j \geqslant 0}$. Thus the function $m^\ast(\lambda)$ is the sum of  functions $m^\ast_1(\lambda) + \cdots + m^\ast_d(\lambda)$.

In terms of norms of functions, we have a norm equal to  
$
\| g \|_\F^2 = \sum_{j=1}^d \| f_j \|_{\F_j}^2.
$

In the particular case where $\mu_{jm_j} \propto m_j^{-2s}$ for all $j$, or equivalently, a number of eigenvalues of $k_j$ greater than $\lambda$ proportional to $\lambda^{-1/(2s)}$,  we have a number of eigenvalues of $k$ greater than $\lambda$ equivalent to $ d \lambda^{-1/(2s)}$, that is a decay for the eigenvalues proportional to $(m/d)^{-2s}$. Similarly, when the decay is exponential as $\exp(- \rho m)$, we get a decay of $\exp( - \rho m / d)$.

\subsection{Product of kernels: $k(x,y) = \prod_{j=1}^d k_j(x_j,y_j)$}

 In this situation, the RKHS for $K$ is exactly the tensor product of $\F_1,\dots,\F_d$, i.e., the span of all functions $\prod_{j=1}^d f_j(x_j)$, for $f_1,\dots,f_d $ in $\F_1,\dots,\F_d$~\citep{berlinet2004reproducing}. Moreover, the integral operator for $k$ is a tensor product of  the $d$ integral operator for $k_1,\dots,k_d$. This implies that its eigenvalues are $\mu_{1 m_1} \times \cdots  \times\mu_{d m_d}$, $m_1,\dots,m_d \geqslant 0$.   
 In terms of norms of functions defined on $\X_1 \times \cdots \times \X_d$, this thus corresponds to 
$$
 \sum_{m_1,\dots,m_d \geqslant 0} 
 \bigg(  \prod_{j=1}^d\mu_{j m_j} \bigg)^{-1} \bigg\langle f, \prod_{j=1}^d e_{j m_j}(x_j) \bigg\rangle^2_{L_2(d \rho^{\otimes d})}.
$$

\paragraph{Special cases.}

In the particular case where $\mu_{jm_j} \propto m_j^{-2s}$ for all $j$,  we have a number of eigenvalues of $k$ greater than $\lambda$ equivalent to the number of multi-indices such that
$m_1 \times \cdots \times m_d $ is less than $\lambda^{-1/(2s)}$. By counting first the index $m_1$, this can be upper-bounded by the sum  of $\frac{\lambda^{-1/(2s)}}{m_2 \cdots m_d}$ over all indices $m_2,\dots, m_d$ less than $\lambda^{-1/(2s)}$, which is less than $\lambda^{-1/(2s)} \big( \sum_{m=1}^{\lambda^{-1/(2s)} }\frac{1}{m} \big)^{d-1} = O \Big( \lambda^{-1/(2s)} 
\big( s \log \frac{1}{\lambda} \big)^{d-1} \Big)$.
 This results in a decay of eigenvalues bounded by $( \log m)^{2s(d-1)} m^{-2s}$ (this can be obtained by inverting approximately the function of $\lambda$).

 When the decay is exponential as $\exp ( - \rho \lambda)$, then we get that $m^\ast(\lambda)$ is the number of multi-indices $(m_1,\dots,m_d)$ such that their sum is less than $c = \frac{\log \frac{1}{\lambda}}{\rho}$; when $c$ is large, this is equivalent to $c^d$ times the volume of the $d$-dimensional simplex, and thus less than $\frac{c^d }{ d!} = \big( \frac{\log \frac{1}{\lambda}}{\rho}\big)^d \frac{1}{d!}$. This leads to a decay of eigenvalues as $\exp( - \rho d!^{1/d} m^{1/d} )$ or, by using Stirling formula, less than $\exp( - \rho  d m^{1/d} )$.

\section{Proofs}

\subsection{Proof of Prop.~\ref{prop:upper}}
\label{app:upper}

As shown in \mysec{kexp}, any $f \in \F$ with  $\F$-norm less than one may be represented as $f = \int_\V g(v) \varphi(v,\cdot) d\tau(v)$, for a certain  $g \in L_2(d\tau)$ with $L_2(d\tau)$-norm less than one. We do not solve the problem in $\beta$ exactly, but use a properly chosen Lagrange multiplier $\lambda$ and consider the following minimization problem:
$$
 \bigg\| 
\sum_{i=1}^n \beta_i q(v_i)^{-1/2} \varphi(v_i,\cdot) - \int_\X \varphi(v,\cdot) g(v) d\tau(v)
\bigg\|_{L_2(d\rho)}^2 +  {n\lambda}  \| \beta\|_2^2.
$$
We consider the operator $\Phi: \rb^n \to L_2(d\rho)$ such that
$$\Phi \beta = \sum_{i=1}^n \beta_i q(v_i)^{-1/2} \varphi(v_i,\cdot). $$
We then need to minimize the familiar least-squares problem:
$$
 \big\| f - \Phi \beta \big \|_{L_2(d\rho)}^2 +  {n\lambda}  \| \beta\|_2^2,
$$
with solution from the usual normal equations and the matrix inversion lemma for operators~\citep{ogawa1988operator}:
\BEQ
\label{eq:beta}
\beta = ( \Phi^\ast \Phi + n \lambda \idm)^{-1} \Phi^\ast f= \frac{1}{n} \Phi^\ast ( \frac{1}{n}\Phi \Phi^\ast +   \lambda \idm)^{-1} f.
\EEQ
We consider the empirical integral operator $\hat{\Sigma}: L_2(d\rho) \to L_2(d\rho)$, defined as
$$
\hat{\Sigma} = \frac{1}{n} \Phi \Phi^\ast = \frac{1}{n}\sum_{i=1}^n \frac{1}{q(v_i)} \varphi(v_i,\cdot)  \otimes_{L_2(d\rho)} \varphi(v_i,\cdot),
$$
that is, for $a,b \in L_2(d\rho)$, 
$ \displaystyle \langle a,\hat{\Sigma} b \rangle_{ L_2(d\rho)} = 
\sum_{i=1}^n \frac{\langle a, \varphi(v_i,\cdot) \rangle_{ L_2(d\rho)}  \langle b, \varphi(v_i,\cdot) \rangle_{ L_2(d\rho)} }{q(v_i)} $. By construction, and following the end of \mysec{kexp}, we have $\E \hat{\Sigma} = \Sigma$.

The value of $\| f - \Phi \beta \|_{ L_2(d\rho)}^2$ is equal to
\BEA
\nonumber \| f - \Phi \beta \|_{ L_2(d\rho)}^2
& = & \| f - \hat{\Sigma}  ( \hat{\Sigma} + \lambda \idm )^{-1} f \|_{ L_2(d\rho)}^2
= \| \lambda ( \hat{\Sigma} + \lambda \idm )^{-1} f \|_{ L_2(d\rho)}^2
\\
\label{eq:AA}& = & \lambda^2 \big\langle f, ( \hat{\Sigma} + \lambda \idm )^{-2} f \big\rangle_{ L_2(d\rho)}
\leqslant \lambda \big\langle f, ( \hat{\Sigma} + \lambda \idm )^{-1} f \big\rangle_{ L_2(d\rho)},
\EEA
because $( \hat{\Sigma} + \lambda \idm )^{-2} \preccurlyeq \lambda^{-1} ( \hat{\Sigma} + \lambda \idm )^{-1}$ (with the classical partial order between self-adjoint operators).

Finally, we have, with $\beta = \frac{1}{n} \Phi^\ast ( \hat{\Sigma} + \lambda \idm)^{-1} f $:
\BEQ
\label{eq:BB}
n\| \beta\|_2^2  = \big \langle ( \hat{\Sigma} + \lambda \idm )^{-1} f ,  \hat{\Sigma}
( \hat{\Sigma} + \lambda \idm )^{-1} f
\big\rangle_{ L_2(d\rho)} 
\leqslant \big \langle  f ,  
( \hat{\Sigma} + \lambda \idm )^{-1} f
\big\rangle_{ L_2(d\rho)} ,
\EEQ
using $( \hat{\Sigma} + \lambda \idm )^{-2} \hat{\Sigma} \preccurlyeq   ( \hat{\Sigma} + \lambda \idm )^{-1}$.

By construction, we have $ \E(\hat{\Sigma}) = \Sigma$. Moreover, we have, by Cauchy-Schwarz inequality:
\BEAS
 \langle a, ( f \otimes_{ L_2(d\rho)} f )  a \rangle_{ L_2(d\rho)} 
 & \!\!\!\!\!= \!\!\!\!\!& \bigg( \int_\X a(x) f(x) d\rho(x)  \bigg)^2
 = \bigg( \int_\X \int_\V a(x) g(v) \varphi(v,x) d\tau(v)  d\rho(x)  \bigg)^2 \\
 &\!\!\!\!\! \leqslant \!\!\! \!\!& 
 \bigg( \int_\V g(v)^2 d\tau(v) \bigg)\!\! \int_\V \!\! \bigg( \int_\X \!\! a(x)  \varphi(v,x)    d\rho(x)  \bigg)^2 d\tau(v) 
\\
& =&  \| g\|^2_{L_2(d\rho)}  \langle a, {\Sigma} a \rangle_{ L_2(d\rho)}  
 \leqslant   \langle a, {\Sigma} a \rangle_{ L_2(d\rho)} .
\EEAS
Thus $f \otimes_{ L_2(d\rho)} f \preccurlyeq \Sigma$, and we may thus define
$\langle f, \Sigma^{-1} f \rangle_{L_2(d \rho)} $, which is less than one.

 Overall we aim to study
 $
   \langle f, ( \hat{\Sigma} + \lambda \idm )^{-1} f \rangle_{L_2(d \rho)}$, 
 for $\langle f, \Sigma^{-1} f \rangle_{L_2(d \rho)} \leqslant 1$, to control both the norm $\|\beta\|_2^2$ in \eq{BB} and the approximation error $\| f - \Phi \beta \|_{ L_2(d\rho)}^2$ in \eq{AA}.
We have, following a similar argument than the one of \citet{bac2012sharp,alaoui2014fast} for column sampling, i.e., by  a formulation using $\Sigma - \hat{\Sigma}$ in terms of operators in an appropriate way:
\BEAS
& &   \langle f, ( \hat{\Sigma} + \lambda \idm )^{-1} f \rangle_{L_2(d \rho)} \\
& =  & \langle f, ( \Sigma  + \lambda \idm + \hat{\Sigma} - \Sigma  )^{-1} f \rangle_{L_2(d \rho)}
\\
& =  &   \big\langle  ( \Sigma  + \lambda \idm)^{-1/2} f, \big[ \idm +   ( \Sigma  + \lambda \idm)^{-1/2} ( \hat{\Sigma} - \Sigma)  ( \Sigma  + \lambda \idm)^{-1/2}    \big]^{-1}  ( \Sigma  + \lambda \idm)^{-1/2} f \big\rangle_{L_2(d \rho)}.
\EEAS
Thus, if $( \Sigma  + \lambda \idm)^{-1/2} ( \hat{\Sigma} - \Sigma)  ( \Sigma  + \lambda \idm)^{-1/2} \succcurlyeq - t \idm$, with $ t\in (0,1)$, we have  
\BEAS
  \langle f, ( \hat{\Sigma} + \lambda \idm )^{-1} f \rangle_{L_2(d \rho)}
& \leqslant &  
 \langle  ( \Sigma  + \lambda \idm)^{-1/2} f,    ( 1-t)^{-1} ( \Sigma  + \lambda \idm)^{-1/2} f \rangle_{L_2(d \rho)}
\\
& = & (1-t)^{-1} 
 \langle f,    ( \Sigma  + \lambda \idm)^{-1} f \rangle_{L_2(d \rho)}
\\
& \leqslant & (1-t)^{-1} 
  \langle f,     \Sigma ^{-1} f  \rangle_{L_2(d \rho)} \leqslant  (1-t)^{-1}  
 .
\EEAS
Moreover, we have shown $ ( \hat{\Sigma} + \lambda \idm )^{-1} \preccurlyeq  \frac{1}{1-t} (  {\Sigma} + \lambda \idm )^{-1}$.

Thus, the performance depends on having 
 $( \Sigma  + \lambda \idm)^{-1/2} ( {\Sigma} - \hat\Sigma)  ( \Sigma  + \lambda \idm)^{-1/2} \preccurlyeq  t \idm$.
 
We consider the  self-adjoint operators $X_i$, for $i=1,\dots,n$, which are independent and identically distributed:
$$X_i = \frac{1}{n} ( \Sigma  + \lambda \idm)^{-1} \Sigma -  \frac{1}{n} \frac{1}{q(v_i) }\big[  ( \Sigma  + \lambda \idm)^{-1/2} \varphi(v_i,\cdot) \big]
 \otimes_{L_2(d\rho)} \big[  ( \Sigma  + \lambda \idm)^{-1/2}  \varphi(v_i,\cdot)\big],$$
 so that our goal is to provide an upperbound on the probability that 
 $
 \| \sum_{i=1}^n X_i \|_{\rm op} > t
 $, where $\|\cdot \|_{\rm op}$ is the operator norm (largest singular values). We use the notation
 $$d = \tr \Sigma (\Sigma + \lambda \idm)^{-1} = \int_\V  \frac{\langle  \varphi(v,\cdot), ( \Sigma  + \lambda \idm)^{-1}   \varphi(v,\cdot)\rangle_{L_2(d\rho)} }{q(v)} q(v) d\tau(v) \leqslant d_{\max}
  .$$
 We have 
 \BEAS
 \E X_i & = & 0, \mbox{ by construction of } X_i, \\
 X_i & \preccurlyeq &  \frac{1}{n} ( \Sigma  + \lambda \idm)^{-1} \Sigma \preccurlyeq 
 \frac{1}{n} \tr \big[ ( \Sigma  + \lambda \idm)^{-1} \Sigma \big] \idm  \preccurlyeq
 \frac{d_{\max}}{n} \idm ,\\
 X_i & \succcurlyeq & -  \frac{1}{n} \frac{1}{q(v_i) }\big[  ( \Sigma  + \lambda \idm)^{-1/2} \varphi(v_i,\cdot) \big]
 \otimes_{L_2(d\rho)} \big[  ( \Sigma  + \lambda \idm)^{-1/2}  \varphi(v_i,\cdot)\big] \\
 & \succcurlyeq & -  \frac{1}{n} \frac{1}{q(v_i) } \big\|  ( \Sigma  + \lambda \idm)^{-1/2} \varphi(v_i,\cdot) \big\|^2_{L_2(d \rho)} \idm  \succcurlyeq -  \frac{d_{\max}}{n} \idm ,  \\
 \| X_i \|_{\rm op}   &  \leqslant &   \frac{d_{\max}}{n} \mbox{ as a consequence of the two previous inequalities,} \\
 \E (X_i^2) & = &  
\E  \bigg[
 \frac{1}{n} \frac{1}{q(v_i) }\big[  ( \Sigma  + \lambda \idm)^{-1/2} \varphi(v_i,\cdot) \big]
 \otimes_{L_2(d\rho)} \big[  ( \Sigma  + \lambda \idm)^{-1/2}  \varphi(v_i,\cdot)\big]
 \bigg]^2
 - \big[  \frac{1}{n} ( \Sigma  + \lambda \idm)^{-1} \Sigma \big]^2
 \\
 & \preccurlyeq & \E \bigg[
 \frac{1}{n} \frac{1}{q(v_i) }\big[  ( \Sigma  + \lambda \idm)^{-1/2} \varphi(v_i,\cdot) \big]
 \otimes_{L_2(d\rho)} \big[  ( \Sigma  + \lambda \idm)^{-1/2}  \varphi(v_i,\cdot)\big]
 \bigg]^2 \\
 & = &   \E
   \frac{\langle  \varphi(v_i,\cdot), ( \Sigma  + \lambda \idm)^{-1}  \varphi(v_i,\cdot) \rangle_{L_2(d\rho)} }{n^2 q(v_i)^2 }\big[  ( \Sigma  + \lambda \idm)^{-1/2} \varphi(v_i,\cdot) \big]
 \otimes_{L_2(d\rho)} \big[  ( \Sigma  + \lambda \idm)^{-1/2}  \varphi(v_i,\cdot)\big]
 \\
 & \preccurlyeq & \frac{d_{\rm max} }{n^2}
 \E \bigg( \big[  \frac{1}{q(v_i)} ( \Sigma  + \lambda \idm)^{-1/2} \varphi(v_i,\cdot) \big]
 \otimes_{L_2(d\rho)} \big[  ( \Sigma  + \lambda \idm)^{-1/2}  \varphi(v_i,\cdot)\big] \bigg)
 = \frac{d_{\rm max} }{n^2} \Sigma ( \Sigma  + \lambda \idm)^{-1} ,
  \\
  \!\!\!\! \!\!\!\!\!\!\sum_{i=1}^n \E ( X_i^2) 
   & \preccurlyeq &  
   \frac{d_{\max}}{n}  (\Sigma+\lambda \idm)^{-1} \Sigma,
   \EEAS
   with a maximal eigenvalue less than $ \displaystyle \frac{d_{\max}}{n}$ and a trace less than 
  $\displaystyle \frac{d_{\max}}{n}  \tr \Sigma(\Sigma +\lambda \idm)^{-1} = \frac{d \, d_{\max}}{n} $.

 Following~\citet{hsu2014random}, we  use a matrix Bernstein inequality which is independent of the underlying dimension (which is here infinite). We consider the bound of~\citet[Theorem 2.1]{minsker2011some}, which improves on the earlier result of~\citet[Theorem 4]{hsu2012tail}, that is:
  $$
 \P \bigg(
    \bigg\| \sum_{i=1}^n X_i \bigg\|_{\rm op} > t\bigg)
 \leqslant 
 2d \bigg( 1 + \frac{ 6   }{t^2 \log^2(1+ nt/d_{\rm max} ) } \bigg) \exp\bigg( - \frac{   t^2 / 2 }{  d_{\max}/n ( 1 + t/3) } \bigg)
 $$
 
 We now consider $t = \frac{3}{4}$, $\delta \in (0,1)$, and
 $ \displaystyle n \geqslant  B d_{\max} \log \frac{ C d_{\max} }{\delta} $, with appropriate constants $B,C>0$. This   implies that
 $$
\exp\bigg( - \frac{   t^2 / 2 }{  d_{\max}/n ( 1 + t/3) } \bigg)
\leqslant \exp\bigg( - \frac{   (3/4)^2 / 2 }{  5/4  }  B \log \frac{ C d_{\max} }{\delta} \bigg)
 \leqslant \big(\frac{\delta}{C d_{\max} }\big)^{\frac{   (3/4)^2 B / 2  }{  5/4  }   }
 \leqslant \big(\frac{\delta}{C d }\big)^{\frac{   (3/4)^2 B / 2  }{  5/4  }   },
 $$
 and, if $d_{\max} \geqslant D$, using $n \geqslant  B d_{\max}  \log CD$,
 $$
  1 + \frac{ 6   }{t^2 \log^2(1+ nt/d_{\rm max} ) }  
 \leqslant 1 + \frac{ 6 \cdot 16 / 9}
{ \log^2 \big( 1 + (3B/4)    \log  (C D)  \big)},
 $$
 while if $d_{\max} \leqslant D$ and $n \geqslant 1$,
$$
  1 + \frac{ 6   }{t^2 \log^2(1+ nt/d_{\rm max} ) }  
 \leqslant 1 + \frac{ 6 \cdot 16 / 9}
{ \log^2 \big( 1 + (3/4D)   \big)}.
 $$
In order to get a bound, we need $\frac{   (3/4)^2 B / 2  }{  5/4  }  \geqslant 1$, and we can take $B=5$. If we take $C=8$, then in order to have $ 1 + \frac{ 6   }{t^2 \log^2(1+ nt/d_{\rm max} ) }   \leqslant 4$, we can take $D = 3/8$. 
 Thus the probability is less than $ \delta$.

Finally, in order to get the extra bound on $ \frac{1}{n} \sum_{i=1}^n q(v_i)^{-1} \|  \varphi(v_i,\cdot)  \|_{L_2(d\rho)}^2$, we consider  $\E \tr \hat{\Sigma} = \tr \Sigma = \int_\X k(x,x) d\rho(x)$, and thus, by Markov's inequality,  with probability $1-\delta$, 
 \BEQ
 \label{eq:bound}
 \frac{1}{n} \sum_{i=1}^n q(v_i)^{-1} \|  \varphi(v_i,\cdot)  \|_{L_2(d\rho)}^2 = \tr \hat{\Sigma}  \leqslant \frac{1}{\delta}\tr \Sigma.
 \EEQ 
 By taking $\delta/2$ instead of $\delta$ in the control of  $
 \| \sum_{i=1}^n X_i \|_{\rm op} > t
 $ and in the Markov inequality above, we have a control over $\| \beta\|_2^2$, $\tr \hat{\Sigma}$ and the approximation error, which leads to the desired result in Prop~\ref{prop:upper}. This will be useful for the lower bound  of Prop.~\ref{prop:lower}.

 We can make the following extra observations regarding the proof:
 \BIT
 \item[--] It may be possible to derive  a similar result with a thresholding of eigenvalues in the spirit of~\citet{zwald2004kernel}, but this would require Bernstein-type concentration inequalities for the projections on principal subspaces.
 \item[--] We have seen that with high-probability, we have
$ ( \hat{\Sigma} + \lambda \idm )^{-1} \preccurlyeq  4 (  {\Sigma} + \lambda \idm )^{-1}$. Note that    $A \preccurlyeq B$ does not imply in  $A^2 \preccurlyeq B^2$~\citep[][page 9]{bhatia2009positive} and that in general we do not have $ (\hat{\Sigma} +\lambda \idm)^{-2}  \preccurlyeq C (\Sigma +\lambda \idm)^{-2} $ for any constant $C$ (which would allow an improvement in the error by replacing $\lambda$ by $\lambda^2$, and violate the lower bound of Prop.~\ref{prop:lower}).
 
 \item[--] We may also obtain a result in expectation, by using $\delta = 4 \lambda / \tr \Sigma$ (which is assumed to be less than 1), leading to a squared error with expectation less than $8 \lambda$ as soon as 
 $n \geqslant 5 d_{\max}(\lambda) \log \frac{  2 (\tr \Sigma) d_{\max} (\lambda)}{\lambda}  .$ 
 Indeed, we can use the bound $4 \lambda$ with probability $1-\delta$ and $ \| f\|_{L_2(d\rho)}^2 \leqslant \tr \Sigma $ with probability $\delta$, leading to a bound of $ 4 \lambda ( 1- \delta) + \delta \tr \Sigma \leqslant 8 \lambda$.
 We use this result in \mysec{consequences}.

  \EIT

\subsection{Proof of Prop.~\ref{prop:upper2}}
\label{app:upper2}
We start from the bound above, with the constraint $n \geqslant 5 d (\lambda) \log \frac{  16 d  (\lambda)}{\delta}  .$ Statement (a) is a simple reformulation of Prop.~\ref{prop:upper}. For statement (b), if we assume
$m \leqslant \frac{ n}{5 (1+ \gamma) \log \frac{16 n}{5 \delta}}$, and $\lambda = \mu_m$, then we have
$d(\lambda) \leqslant (1+\gamma) m$, which implies
 $n \geqslant 5 d (\lambda) \log \frac{  16 d  (\lambda)}{\delta} $, and (b) is a consequence of (a).

\subsection{Proof of Prop.~\ref{prop:lower}}
\label{app:lower}

We first use the Varshamov-Gilbert's lemma~\citep[see, e.g.,][Lemma 4.7]{massart-concentration}. That is, for any integer $s$, there exists a family $(\theta_j)_{j \in J}$ of at least $|J| \geqslant e^{s/8}$ distinct elements of $\{0,1\}^s$, such that for $j\neq j' \in J$, $\| \theta_j - \theta_{j'} \|_2^2 \geqslant \frac{s}{4}$.

For each $\theta \in \{0,1\}^s$, we define an element of $\F$ with  norm less than one, as
$f(\theta) = \frac{\sqrt{\mu_{s}}}{\sqrt{s}} \sum_{i=1}^s \theta_i  e_i \in \F$, where $(e_{i},\mu_{i})$, $i=1,\dots,s$ are the eigenvector/eigenvalue pairs associated with the $s$ largest eigenvalues of $\Sigma$. We  have,
since $\mu_i \geqslant \mu_s$ for $i \in \{1,\dots,s\}$ and $\|\theta\|_2^2 \leqslant 1$:
$$
\| f(\theta) \|_\F^2 = \frac{\mu_{s}}{s} \sum_{i=1}^s \theta_i^2 \mu_{i}^{-1} 
\leqslant \frac{\mu_s}{s} \sum_{i=1}^s \theta_i^2 \mu_{s}^{-1} \leqslant \frac{1}{s} \sum_{i=1}^s \theta_i^2 \leqslant 1.
$$
Moreover, for any $j\neq j' \in J$, we have
$
\| f(\theta_j) - f(\theta_{j'})\|_{L_2(d\rho)} ^2 =\frac{\mu_{s}}{s}   \| \theta_j - \theta_{j'} \|_2^2 \geqslant \frac{\mu_{s}}{4}$.

We now assume that $s$ is selected so that $\sqrt{4 \lambda } \leqslant \sqrt{ \frac{\mu_{s}}{4} }/ 3$. 
By applying the existence results to all functions $f_j$, $j \in J$, then there exists a 
 family $(\beta_j)_{j \in J}$ of elements of $\rb^n$, with squared $\ell_2$-norm less than $\frac{4}{n}$, and for which,
 for all $j$,
 $$
 \Big\| f_j - \sum_{i=1}^n (\beta_j)_i \psi_i \Big\|_{L_2(d\rho)} \leqslant \sqrt{4 \lambda}.
 $$
 This leads to,
for any $j\neq j' \in J$, 
\BEAS
\Big\| \sum_{i=1}^n (\beta_j - \beta_{j'})_i \psi_i \Big\|_{L_2(d\rho)}
& \!\!\!\!\geqslant \!\!\!\!& \big\| f_j - f_{j'}\big\|_{L_2(d\rho)}  - \Big\| \sum_{i=1}^n (\beta_j )_i \psi_i  - f_j\Big\|_{L_2(d\rho)}
- \Big\| \sum_{i=1}^n (\beta_{j'} )_i \psi_i  - f_{j'}\Big\|_{L_2(d\rho)}  \\
&  \!\!\!\!\geqslant  \!\!\!\!& \sqrt{  \mu_s / 4} - 2 \sqrt{ \frac{\mu_{s}}{4}}/3= \sqrt{ \frac{\mu_{s}}{4}}/3.
\EEAS
Moreover, we have the bound
$$
\bigg\| \sum_{i=1}^n (\beta_j - \beta_{j'})_i \psi_i \bigg\|_{L_2(d\rho)}^2
\leqslant \bigg( \sum_{i=1}^n  (\beta_j - \beta_{j'})_i^2 \bigg)
\sum_{i=1}^n \| \psi_i  \|_{L_2(d\rho)}^2
\leqslant \| \beta_j - \beta_{j'}\|_2^2  \ \cdot n
( 2 \delta^{-1} \tr \Sigma ) .
$$
Combining the last two inequalities, we get 
$\| \beta_j - \beta_{j'} \|_2 \geqslant \sqrt{ \frac{\delta \mu_{s}}{72 n  \tr \Sigma}} = \Delta $. Thus,  $e^{s/8}$
is less than the $\Delta$-packing number of the ball of radius $r = 2 / \sqrt{n}$, which is itself less than
$(r/\Delta)^n ( 2 + \Delta/r)^n$~\citep[see, e.g.,][Lemma 4.14]{massart-concentration}. Since
$\Delta/r  = \sqrt{ \frac{\delta \mu_{s}}{4 \cdot 72    \tr \Sigma}} 
\leqslant \frac{1}{12 \sqrt{2}}$,   we have 
$$
\frac{s}{8} \leqslant n \bigg( \frac{1}{2}\log \frac{4 \cdot 72  \tr \Sigma}{\delta \mu_{s}}
+ \log ( 2 +  \frac{1}{12 \sqrt{2}}) \bigg).
$$
This implies  
$
n   \geqslant   \frac{s}{ 
4 \log \frac{  \tr \Sigma}{\delta \mu_{s}}
+ 29 }.
$
Given that we have to choose $\mu_{s} \geqslant 144 \lambda$ for the result to hold, this implies the desired result, since $4 \log(1440) \geqslant 29$.

\end{document}